\documentclass{article}

\usepackage[final]{neurips_data_2024}

\usepackage{natbib}
\setcitestyle{numbers,square,sort}

\usepackage[utf8]{inputenc} %
\usepackage[T1]{fontenc}    %
\usepackage{hyperref}       %
\usepackage{url}            %
\usepackage{booktabs}       %
\usepackage{amsfonts}       %
\usepackage{nicefrac}       %
\usepackage{microtype}      %
\usepackage[table, svgnames, dvipsnames]{xcolor}
\usepackage{cellspace}
\usepackage{tocloft}

\usepackage{amsmath}
\usepackage{amssymb}
\usepackage{bm}
\usepackage{bbm}
\usepackage{xspace}
\usepackage{cleveref}
\usepackage{subcaption} %
\usepackage{multirow} 
\usepackage{wrapfig}
\usepackage{minitoc}
\usepackage[framemethod=tikz]{mdframed}
\usepackage[final]{pdfpages} 
\usepackage{longtable}

\usepackage{rotating}

\usepackage{mdframed}

\usepackage{pifont}

\newcommand{\inquire}{\textsc{Inquire}}

\newcommand*{\ie}{i.e.,\@\xspace}
\newcommand*{\eg}{e.g.,\@\xspace}
\newcommand*{\etc}{etc.\@\xspace}

\newcommand{\cmark}{\ding{51}}%
\newcommand{\xmark}{\ding{55}}%

\usepackage{graphicx}

\title{}

\title{ 
    \inquire{}: A Natural World \\Text-to-Image Retrieval Benchmark
}

\author{
  \textbf{Edward Vendrow\thanks{Equal contribution. \textdagger Equal supervision, order randomized.}}\ \ \textsuperscript{1}, 
  \textbf{Omiros Pantazis\footnotemark[1]}\ \ \textsuperscript{2}, 
  \textbf{Alexander Shepard\textsuperscript{3}},
  \textbf{Gabriel Brostow\textsuperscript{2}}, \\
  \textbf{Kate E.~Jones\textsuperscript{2}},
  \textbf{Oisin Mac Aodha\textsuperscript{\textdagger~4}},
  \textbf{Sara Beery\textsuperscript{\textdagger~1}},
  \textbf{Grant Van Horn\textsuperscript{\textdagger~5}}\\
  \\
  \textsuperscript{1}Massachusetts Institute of Technology \quad
  \textsuperscript{2}University College London \quad
  \textsuperscript{3}iNaturalist \quad \\
  \textsuperscript{4}University of Edinburgh \quad
  \textsuperscript{5}University of Massachusetts Amherst
}

\begin{document}

\maketitle

\begin{figure}[ht]
    \centering
    \vspace{-6pt}
    \includegraphics[width=\textwidth]{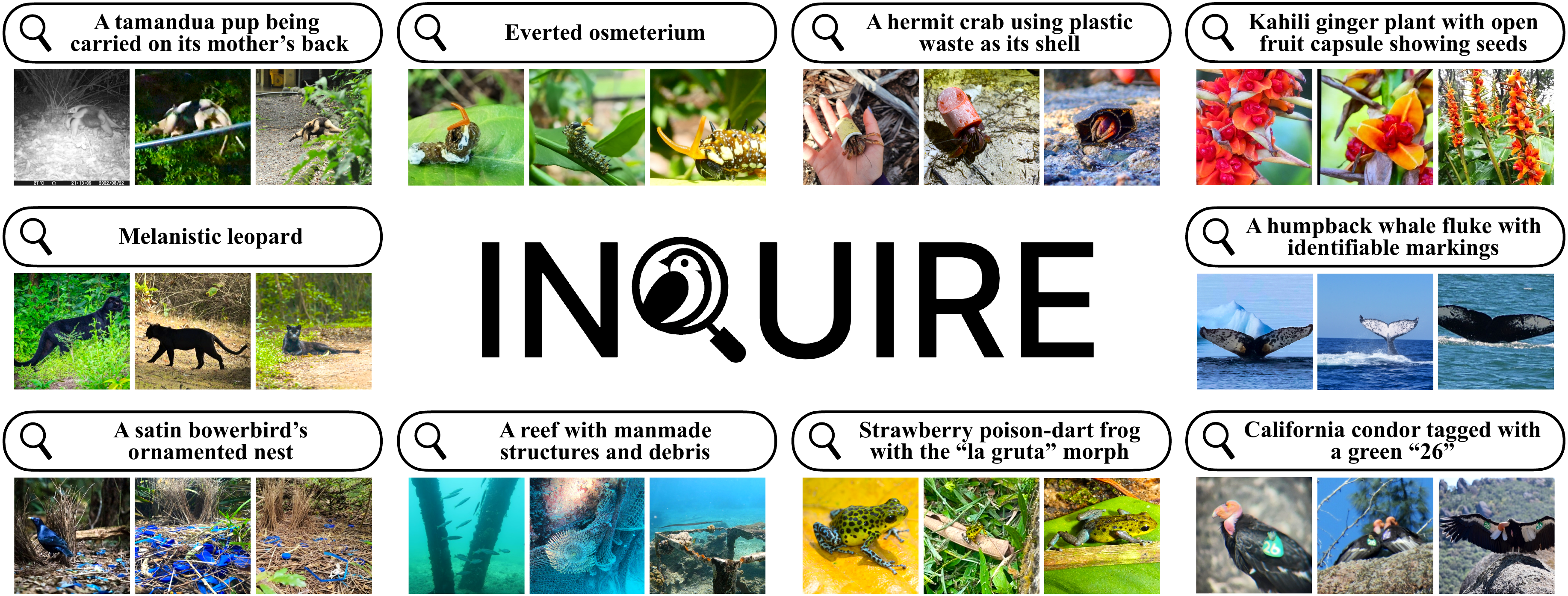}
    \caption{\inquire{} is a text-to-image retrieval benchmark of 250 expert-level queries comprehensively labeled over a new five million image dataset. The queries span a range of ecological and biodiversity concepts, requiring reasoning, image understanding, and domain expertise.}
    \label{fig:abstract_teaser}
    \vspace{-2pt}
\end{figure}

\begin{abstract}
We introduce \inquire{}, a text-to-image retrieval benchmark designed to challenge multimodal vision-language models on expert-level queries. 
\inquire{} includes iNaturalist 2024 (iNat24), a new dataset of five million natural world images, along with 250 expert-level retrieval queries. These queries are paired with all relevant images comprehensively labeled within iNat24, comprising 33,000 total matches.
Queries span categories such as species identification, context, behavior, and appearance, emphasizing tasks that require nuanced image understanding and domain expertise.
Our benchmark evaluates two core retrieval tasks: (1) \textsc{Inquire-Fullrank}, a full dataset ranking task, and (2) \textsc{Inquire-Rerank}, a reranking task for refining top-100 retrievals. 
Detailed evaluation of a range of recent multimodal models demonstrates that \inquire{} poses a significant challenge, with the best models failing to achieve an mAP@50 above 50\%. 
In addition, we show that reranking with more powerful multimodal models can enhance retrieval performance, yet there remains a significant margin for improvement.
By focusing on scientifically-motivated ecological challenges, \inquire{} aims to bridge the gap between AI capabilities and the needs of real-world scientific inquiry, encouraging the development of retrieval systems that can assist with accelerating ecological and biodiversity research.
\end{abstract}

\addtocontents{toc}{\protect\setcounter{tocdepth}{0}}
\section{Introduction}

Recent advances in multimodal learning have resulted in advanced models~\cite{radford2021learning,liu2023llava,achiam2023gpt} that demonstrate remarkable generalization capabilities in zero-shot classification~\cite{radford2021learning,zhai2023sigmoid}, visual question-answering (VQA)~\cite{li2022blip,yu2022coca,alayrac2022flamingo, li2023blip}, and image retrieval~\cite{yu2022coca,li2023blip}. 
These models offer the potential to assist in the exploration, organization, and extraction of knowledge from large image collections. 
However, despite this success, there remains a significant gap in the evaluation of these models on domain-specific, expert-level queries, where nuanced understanding and precise retrieval are critical. 
Addressing this gap is essential for future deployment in specialized fields such as biodiversity monitoring and biomedical imaging, among other scientific disciplines. %

Previous studies of the multimodal capabilities of this new generation of models have primarily focused on the task of VQA. 
In VQA, it has been demonstrated that there remains a large performance gap between state-of-the-art models and human experts in the context of challenging perception and reasoning queries such as those found on college-level exams~\cite{yue2023mmmu,zhong2023agieval}. 
However, no such expert-level benchmark exists for \emph{image retrieval}. 
The most commonly used text-to-image retrieval benchmarks are derived from image captioning datasets, and contain simple queries related to common everyday categories~\cite{young2014image,lin2014microsoft}. 
Current multimodal models achieve near perfect performance on some of these benchmarks, indicating that they no longer pose a challenge (\eg BLIP-2~\cite{li2023blip} scores 98.9 on Flickr30K~\cite{young2014image} top-10).
Existing retrieval datasets are generally small~\cite{philbin2007object,philbin2008lost,young2014image,lin2014microsoft}, limited to a single visual reasoning task (\eg landmark-location matching~\cite{philbin2007object,philbin2008lost,weyand2020google}), and lack 
concepts that would require expert knowledge~\cite{philbin2007object,philbin2008lost,weyand2020google, young2014image,lin2014microsoft}. 
These limitations impede our ability to track and improve image retrieval capabilities.

\begin{figure}
    \centering
    \includegraphics[width=0.9\linewidth]{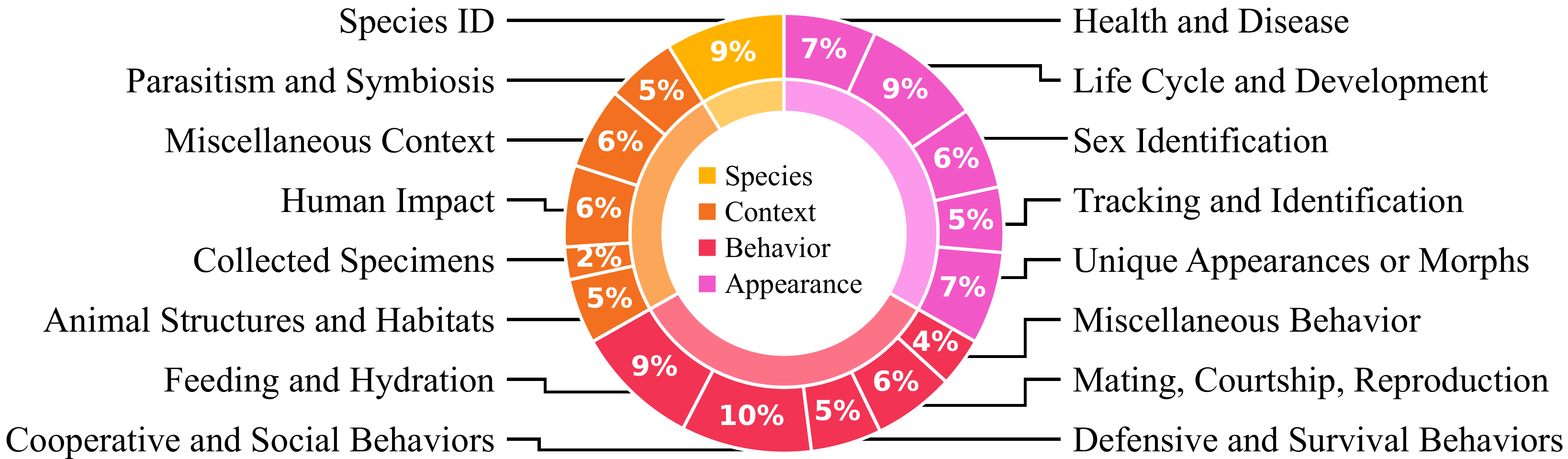}
    \caption{Category breakdown for the fine-grained queries that make up \inquire{}. Each query category falls under one of the following supercategories: Species, Context, Behavior, or Appearance.}    \label{fig:task_by_category}
    \vspace{-3pt}
\end{figure}

A domain that is well-suited for studying this problem is the natural world, where images collected by enthusiast volunteers provide vast and largely uncurated sources of publicly available scientific data. 
In particular, the iNaturalist~\cite{iNaturalist} platform contains over 180 million species images and contributes immensely to research in biodiversity monitoring \cite{chandler2017contribution,Lohan_2024}. %
These images also contain a wealth of ``secondary data'' not reflected in their species labels~\cite{pernat2024overcoming}, including crucial insights into interactions, behavior, morphology, and habitat that could be uncovered  through searches.
However, the time-consuming and expert-dependent analysis needed to extract such information prevents scientists from taking advantage of this valuable data at scale. 
This cost is amplified as scientists typically want to retrieve \emph{multiple} relevant images for each text query, so that they can track changes of a property over space and time~\cite{young2019finding}.
This domain serves as an ideal testbed for expert image retrieval, as these images contain expert-level diverse and composite visual reasoning problems, and progress in this field will enhance impactful scientific discovery. %

In this work, we introduce \inquire{}, a new dataset and benchmark for expert-level text-to-image retrieval and reranking on natural world images. 
\inquire{} includes the iNat24 dataset and 250 ecologically motivated retrieval queries. The queries span 33,000 true-positive matches, pairing each text query with all relevant images that we comprehensively labeled among iNat24's five million natural world images. 
iNat24 is sampled from iNaturalist~\cite{iNaturalist}, and contains images from 10,000 different species collected and annotated by citizen scientists, providing significantly more data for researchers interested in fine-grained species classification.
The queries contained within \inquire{} come from discussions and interviews with a range of experts including ecologists, biologists, ornithologists, entomologists, oceanographers, and forestry experts.

Our evaluation of multimodal retrieval methods demonstrates that \inquire{} poses a significant challenge, necessitating the development of models able to perform expert-level retrieval within large image collections. 
A key finding from our experiments is that reranking, a technique  typically used in text retrieval~\cite{nogueira2019passage,khattab2020colbert,karpukhin2020dense}, offers a promising avenue for improvement in image retrieval.
We hope that \inquire{} will inspire the community to build next-generation image retrieval methods towards the ultimate goal of accelerating scientific discovery.
We make \inquire{}, the iNat24 dataset, pre-computed outputs from state-of-the-art models, and code for evaluation available at \textcolor{blue}{\url{https://inquire-benchmark.github.io/}}.

\begin{figure}
\centering
\begin{minipage}{.48\textwidth}
  \centering
  \includegraphics[width=\textwidth]{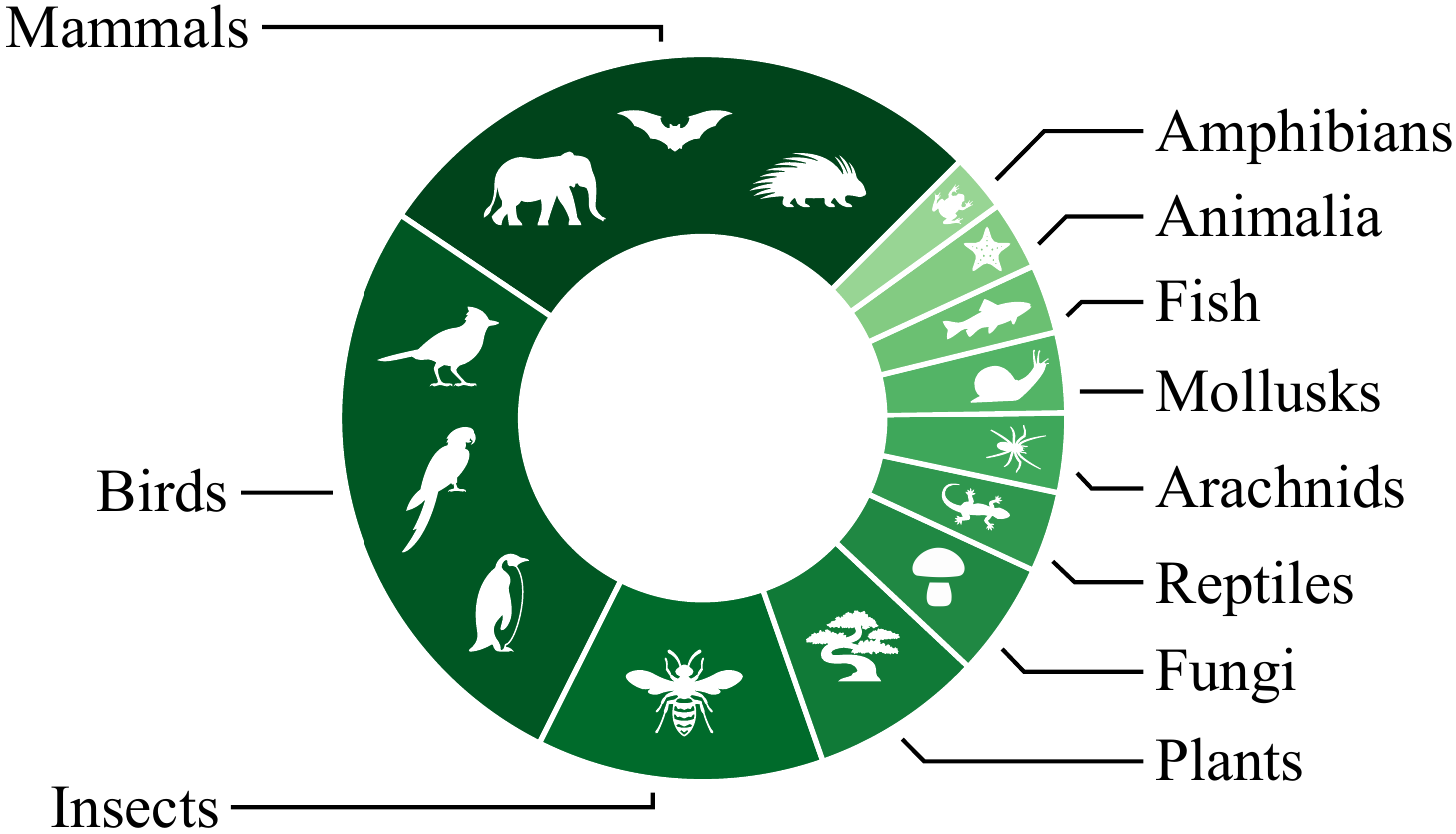}
  \vspace{-12pt}
  \captionof{figure}{Proportion of queries in \inquire{} associated with each iconic group of species.}
  \label{fig:task_by_organism}
\end{minipage}%
\hfill
\begin{minipage}{.48\textwidth}

\centering

    \renewcommand{\arraystretch}{1.1}
    \captionof{table}{Comparison to common datasets used to evaluate text-to-image retrieval~\cite{gadre2023datacomp}. Unlike other datasets, \inquire{} has significantly more images and many matches per query rather than exactly one. \textit{MpQ: Matches per query}%
    }
    
    \small
    \setlength{\tabcolsep}{0.26em}
    \begin{tabular}{lcccc}
        \toprule
        Dataset & Images & Queries & MpQ & Expert  \\
        \midrule
        Flickr30k~\cite{young2014image} & 1,000 & 5k & 1 & \xmark \\
        COCO~\cite{lin2014microsoft} & 5,000 & 25k & 1 & \xmark \\
        \midrule
        \textbf{\inquire{}} & 5,000,000 & 250 & 1--1.5k & \cmark \\
        \bottomrule
    \end{tabular}
    \label{tab:dataset_summary}
\end{minipage}
\end{figure}

\section{Related Work} 
\textbf{Vision-Language Models (VLMs).} 
Large web-sourced datasets containing paired text and images have enabled recent advances in powerful VLMs~\cite{bordes2024introduction,zong2024self}. 
Contrastive methods such as CLIP~\cite{radford2021learning} and ALIGN~\cite{jia2021scaling}, among others, learn an embedding space where the data from the two modalities can be encoded jointly. 
The ability to reason using natural language and images together has yielded impressive results in a variety of text-based visual tasks such as zero-shot classification~\cite{radford2021learning,zhai2023sigmoid}, image captioning~\cite{li2022blip,yu2022coca,alayrac2022flamingo,hu2022scaling, li2023blip}, and text-to-image generation~\cite{nichol2022glide,avrahami2022blended,ramesh2022hierarchical, saharia2022photorealistic,avrahami2023spatext}. 
However, the effectiveness of these contrastive VLMs for more complex compositional reasoning is bottlenecked by the information loss induced by their text encoders~\cite{kamath2023text}. 

There also exists a family of more computationally expensive VLMs that connect the outputs of visual encoders directly into language models. 
Models like LLaVA~\cite{liu2023llava,liu2023improved}, BLIP~\cite{li2022blip,li2023blip,instructblip}, and GPT-4o~\cite{achiam2023gpt,OpenAI2024} have demonstrated impressive vision-language understanding. 
However, despite their potential for answering complex vision-language queries, these models are not suitable for processing large sets of images at interactive rates, which is essential for retrieval, due to their large computational requirements during inference. 
In this paper, we do not introduce new VLMs, but aim to better understand the capabilities and shortfalls of existing methods for text-to-image retrieval. %

\textbf{Image Retrieval.} 
Effective feature representations are essential for achieving strong image retrieval performance. 
Earlier approaches from image-to-image used hand-crafted features~\cite{lowe2004distinctive,bay2006surf} but these have largely been replaced with deep learning-based alternatives~\cite{krizhevsky2012imagenet, babenko2014neural,arandjelovic2016netvlad,balntas2016learning}. 
More recently, in the context of text-to-image retrieval, we have seen the adoption of contrastive VLMs~\cite{radford2021learning,jia2021scaling} trained on web-sourced paired text and image datasets. 
These models enable zero-shot text-based retrieval and have been demonstrated to exhibit desirable scaling properties as training sets become larger~\cite{gadre2023datacomp,fang2023data}. 
However, despite the potential of VLMs for image retrieval, their evaluation has been mostly limited to small datasets adapted from existing image captioning benchmarks, such as Flickr30k~\cite{young2014image} and COCO~\cite{lin2014microsoft}, which contain just 1,000 and 5,000 images, respectively. %
Furthermore, recent models are saturating performance on these less challenging datasets, \eg 
BLIP-2~\cite{li2023blip} scores 98.9 on Flickr30K and 92.6 on COCO top-10 text-to-image retrieval. 
As most text-to-image benchmarks have been derived from image captioning datasets, each query is a descriptive caption that matches exactly one image. %
In contrast, real-world retrievals often involve multiple images relevant to a single query, and the query itself typically does not describe every aspect of the images as thoroughly as a caption does. We compare \inquire{} to common text-to-image retrieval datasets in Table~\ref{tab:dataset_summary}. 

More recent datasets have been purpose-built to probe specific weaknesses of retrieval systems, such as compositionality~\cite{ma2023crepe,hsieh2023sugarcrepe,ray2024cola}, object relationships~\cite{yuksekgonul2022and}, negation~\cite{wang2023clipn,singh2024learn}, and semantic granularity~\cite{xu2024benchmarking}. 
\cite{wildCLIP2024} created a retrieval dataset for camera trap images, but use image captions that were automatically generated from a small set of discrete image attributes, limiting their utility beyond this set. 
The problem of fine-grained retrieval, where there may only be subtle visual differences between concepts of interest, has also been explored extensively~\cite{wei2021fine}.  
However, typically these datasets convert existing classification datasets to the retrieval setting, resulting in small image pools and limited query diversity.  
\inquire{} addresses these shortcomings with a considerably larger image set and fine-grained queries that require advanced image understanding and domain expertise. 

\textbf{Reranking.}
In text retrieval, a common workflow is to first efficiently obtain an initial ranking of documents using pre-computed text embeddings and then \emph{rerank} the top retrievals with a more costly but sophisticated model~\cite{nogueira2019passage,khattab2020colbert,karpukhin2020dense}. 
While VLMs like CLIP~\cite{radford2021learning} enable efficient image retrieval and more expensive models such as GPT-4o~\cite{OpenAI2024} could perform more complex ranking, this workflow has not been extensively explored in text-to-image applications primarily due to a lack of evaluation datasets. 
To this end, \inquire{} introduces a reranking challenge to drive further progress on this task.

\begin{figure}[t]
    \centering
    \includegraphics[width=0.95\textwidth]{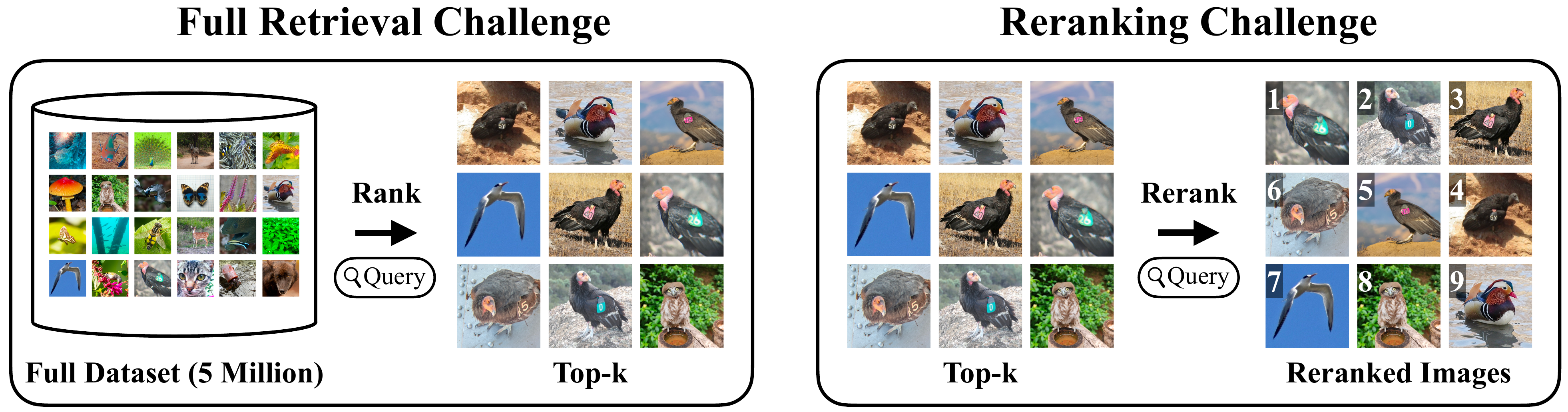}
    \caption{The \inquire{} benchmark consists of a full-dataset ranking task and a reranking task targeting different aspects of the image retrieval problem.}
    \label{fig:challenges}
    \vspace{-5pt}
\end{figure}

\textbf{Expert-level Benchmarks.} 
Visual classification benchmarks have evolved from simply containing common everyday categories~\cite{russakovsky2015imagenet,lin2014microsoft} to having more ``expert-level'' concepts~\cite{van2015building,van2018inaturalist}. 
Challenging datasets like the iNaturalist benchmarks~\cite{van2018inaturalist,van2021benchmarking}, contain large class imbalances and fine-grained concepts that require expert-level knowledge to identify. 
The NeWT benchmark from~\cite{van2021benchmarking} is similar in spirit to \inquire{} in that it proposes a collection of natural world questions. %
However, NeWT is a set of binary classification challenges, and while there are a variety of tasks, the majority of them are standard species classification. Further, NeWT uses a small (200--400) fixed set of positive and negative labeled images for each task, so it is not suitable for evaluating retrieval.

In general, evaluation benchmarks have struggled to keep pace with the growing capabilities of recent large models which perform very well on them~\cite{
,antol2015vqa}. 
For language, specific datasets have been developed to challenge common sense reasoning abilities~\cite{mialon2023gaia,boratko2020protoqa}. 
Multimodal datasets have also been proposed to assess vision-language capabilities~\cite{xu2023lvlm,li2023seed,liu2023mmbench,yin2024lamm}. 
Nevertheless, these benchmarks test general skills in tasks that are not particularly challenging for humans and thus, are not testing a models' abilities in scenarios where expert-level knowledge is required.

To address the need for more difficult benchmarks, recent expert-level benchmarks have been devised for LLMs~\cite{hendrycks2020measuring,zhong2023agieval} and multimodal models~\cite{yue2023mmmu,mensink2023encyclopedic}. 
For instance, MMMU~\cite{yue2023mmmu} features questions that cover a range of college-level disciplines while Encyclopedic-VQA~\cite{mensink2023encyclopedic} comprises visual questions related to fine-grained entities which demand encyclopedic knowledge. %
The relatively low performance on these benchmarks, compared to human performance, highlights current limitations in multimodal models. 
However, there is no equivalent expert-level dataset for fine-grained text-to-image retrieval.  
\inquire{} fills this gap by providing a set of challenging and visually fine-grained retrieval questions focused on real-world tasks in retrieval from natural world image collections.

\section{The \inquire{} Benchmark}\label{sec:benchmark_description}
\vspace{-5pt}
Here we describe \inquire{}, our novel benchmark for assessing expert-level image retrieval for fine-grained queries on natural world image collections. 
\inquire{} consists of a collection of 250 queries, where each query is represented as a brief text description of the concept of interest (\eg \textit{``Alligator lizards mating"~\cite{Pauly_2018}}), and contains its relevant image matches comprehensively labeled over a dataset of five million natural world images. 
These queries represent real scientific use cases collected to cover diverse, expert sources including discussions with scientists across environmental and ecological disciplines. %
Several examples of our queries are illustrated in Figure~\ref{fig:abstract_teaser}, with more in Appendix~\ref{appendix-additional-examples}.
Our queries challenge retrieval methods to demonstrate fine-grained detail recognition, compositional reasoning, character recognition, scene understanding, or natural world domain knowledge. 
While queries can require expert-level knowledge, the information needed to solve them is publicly available online and thus feasible for large web-trained models to learn.
In this section, we detail the data sources utilized for the construction of \inquire{}, describe the data collection process, and introduce two image retrieval tasks --- \textsc{Inquire-Fullrank} and \textsc{Inquire-Rerank} --- that address different aspects of real-world text-to-image retrieval.

\subsection{The iNaturalist 2024 Dataset}
\vspace{-4pt}
As part of the \inquire{} benchmark, we create a new image dataset, which we refer to as iNaturalist 2024 (iNat24). 
This dataset contains five million images spanning 10,000 species classes collected and annotated by community scientists from 2021--2024 on the iNaturalist platform~\cite{iNaturalist}. 
iNat24 forms one of the largest publicly available natural world image repositories, with twice as many images as in iNat21~\cite{van2021benchmarking}. 
To ensure cross-compatibility for researchers interested in using both datasets, iNat24 and iNat21 have the same classes but do not contain the same images, freeing iNat21 to be used as a training set. 
The sampling and collection process of iNat24 is in Appendix~\ref{appendix-data-annotation}.

\subsection{Query and Image Collection Process}
\vspace{-4pt}
\textbf{Query Collection. } To ensure that \inquire{} comprises text queries that are relevant to scientists, we conducted interviews with individuals across different ecological and environmental domains - including experts in ornithology, marine biology, entomology, and forestry. 
Further queries were sourced from reviews of academic literature in ecology~\cite{pernat2024overcoming}.
Representative queries and statistics can be seen in  Figures~\ref{fig:abstract_teaser}, ~\ref{fig:task_by_category}, and~\ref{fig:task_by_organism}. 
We retained only queries that (1) could be discerned from images alone, (2) were feasible to comprehensively label over the entire iNat24 dataset, and (3) were of interest to domain experts. 

\textbf{Image Annotation. } All image annotations were performed by a small set of individuals whose interest and familiarity with wildlife image collections enabled them to provide accurate labels for challenging queries. 
Annotators were instructed to label all candidate images as either \textit{relevant} (\ie positive match) or \textit{not relevant} (\ie negative match) to a query, and to mark an image as not relevant if there was reasonable doubt. 
To allow for comprehensive labeling, where applicable, iNat24 species labels were used to narrow down the search to a sufficiently small size to label all relevant images for the query of interest. 
For queries in which species labels could not be used, labeling was performed over the top CLIP ViT-H-14~\cite{fang2023data} retrievals alone. In this case, the resulting annotations were only kept if we were certain that this labeling captured the vast majority of positives, including labeling until at least 100 consecutive retrievals were not relevant (see Appendix~\ref{appendix-data-annotation}).
Queries that were deemed too easy, not comprehensively labeled, or otherwise not possible to label were excluded from our benchmark. 
In total, this process resulted in 250 queries which involved labeling 194,334 images, of which 32,696 were relevant to their query. Further details are in Appendix~\ref{appendix-data-annotation}.

\textbf{Query Categories. } Each query belongs to one of four supercategories (appearance, behavior, context, or species), and further into one of sixteen fine-grained categories (e.g., Animal Structures and Habitats). Figure~\ref{fig:task_by_category} shows the distribution of query categories, and Figure~\ref{fig:task_by_organism} shows the distribution of iconic groups of the species represented by each query (e.g., Mammals, Birds). We also note queries that use scientific terminology, words typically used only within scientific contexts (e.g., ``A godwit performing distal rhynchokinesis'').

\textbf{Data Split. } We divide all queries into 50 validation and 200 test queries using a random split, stratified by category.

\subsection{Retrieval Tasks}
\vspace{-4pt}
We introduce two tasks to address different aspects of the text-to-image retrieval problem. 
Real-world retrieval implementations often consist of two stages: an initial top-k retrieval with a more computationally efficient method (\eg CLIP zero-shot using pre-computed image embeddings), followed by a reranking of the top-k retrievals with a more expensive model.  
To enable researchers to explore both stages, while ensuring that those with more limited computational resources can participate, we follow previous large-scale reranking challenges like TREC~\cite{craswell2020overview, craswell2021overview} by offering both a full dataset retrieval task and a reranking task (see Figure~\ref{fig:challenges}).

\noindent
\textbf{\textsc{Inquire-Fullrank}. } The goal of this task is end-to-end retrieval, starting from the entire five million image iNat24 dataset. 
Progress on the full retrieval task can be made with better and more efficient ways to organize, process, filter, and search large image datasets. 
Although performance will increase with improvements to either of the two stages in a typical retrieval pipeline, we hope this task also encourages the development of retrieval systems beyond the two-stage approach. %

\textbf{\textsc{Inquire-Rerank}. } This task evaluates reranking performance from a fixed initial ranking of 100 images. 
We believe that significant progress in retrieval will come from developing better reranking methods that re-order an initial retrieved subset. %
Thus, fixing the starting images for each query provides a consistent evaluation of reranking methods. 
This task also lowers the barrier to entry by giving researchers a considerably smaller set of top retrievals to work with, rather than requiring them to implement an end-to-end retrieval system. 
The top 100 ranked images for each query are retrieved using CLIP ViT-H-14 zero-shot retrieval on the entire iNat24 dataset. 
Consistent with previous large-scale reranking challenges~\cite{craswell2020overview, craswell2021overview, lawrie2024overview}, we retain only queries for which at least one positive image is among the top 100 retrieved images and no more than 50\% of these top images are relevant. This ensures that the reranking evaluation remains meaningful and discriminative. 
This filtering process yields a task subset of 200 queries (reduced from our original 250 queries), split into 40 validation and 160 test queries according to the original validation/test split, with and 4,000 and 16,000 corresponding images, respectively.

\vspace{-5pt}
\section{Retrieval Methods} \label{section-methods}
\vspace{-5pt}

The goal of text-to-image retrieval is to rank images from a potentially  large image collection according to their relevance to an input text query. %
Here, we describe the retrieval and reranking methods that we evaluate, covering current state-of-the-art approaches.

\begin{table}
\centering
\caption{\textsc{Inquire-Fullrank} retrieval performance for selected CLIP-style models. Larger models, trained on higher quality datasets, tend to achieve better performance.}
\vspace{0.5em}
\small
\setlength{\tabcolsep}{0.5em}
\renewcommand{\arraystretch}{1}
\begin{tabular}{llcccc}
\toprule

\textbf{Training dataset} & \textbf{Method} & \textbf{Params (M)} & \textbf{mAP@50} & \textbf{nDCG@50} & \textbf{MRR} \\ \midrule

WildCLIP~\cite{wildCLIP2024}                 & CLIP ViT-B-16 & 150 & 7.4  &  16.1  &  0.33 \\
    \midrule
BioCLIP~\cite{stevens2024bioclip}                  & CLIP ViT-B-16 & 150 &  5.0  &  8.6  &  0.17 \\
    \midrule
 \multirow{4}{*}{OpenAI~\cite{radford2021learning}} & CLIP RN50 & 102 & 6.8  &  15.1  &  0.29 \\
                         & CLIP RN50x16 & 291 & 13.6  &  25.5  &  0.48 \\
                         & CLIP ViT-B-32 & 151 & 7.5  &  16.8  &  0.30 \\
                         & CLIP ViT-B-16 & 150 & 10.4  &  20.9  &  0.40 \\
                         & CLIP ViT-L-14 & 428 & 14.4  &  27.1  &  0.46 \\
    \midrule
\multirow{3}{*}{DFN~\cite{fang2023data}}     & CLIP ViT-B-16 & 150 & 15.1  &  28.1  &  0.48 \\
                         & CLIP ViT-L-14 & 428 & 23.1  &  37.3  &  0.54 \\
                         & CLIP ViT-H-14@378 & 987 & \underline{33.3}  &  \underline{48.8}  &  \textbf{0.69} \\
    \midrule
\multirow{2}{*}{WebLI~\cite{zhai2023sigmoid}}   & SigLIP ViT-L-16@384 & 652 & 31.1  &  46.6  &  0.68 \\
                         & SigLIP SO400m-14@384 & 878 & \textbf{34.2}  &  \textbf{49.1}  &  \textbf{0.69} \\
\bottomrule
\end{tabular}
\label{tab:zero_shot_results}
\vspace{-6pt}
\end{table}

\noindent{\bf Embedding Similarity.}
Models such as CLIP~\cite{radford2021learning} are well suited for the text-to-image retrieval setting as they operate on a joint vision and language embedding space. 
In this setting, similarity between an image and text query is simply determined by their cosine similarity. %
The key advantage of embedding models is that the embedding for each image  can be pre-computed once offline as they do not change over time. 
At inference time, only the embedding of the text query needs to be computed and then compared to the cached image embeddings for retrieval.
This is helpful as the number of images we wish to search over can be on the order of millions, or even billions~\cite{schuhmann2022laion}. 
Thus to speed up retrieval, the image embeddings can be pre-computed and indexed using approximate nearest neighbor methods~\cite{douze2024faiss}, allowing for near-instantaneous retrievals on large collections. 
This is beneficial both for end-to-end retrieval and as the first step for a multi-stage retrieval approach.
We also benchmark recent models such as WildCLIP~\cite{wildCLIP2024} and BioCLIP~\cite{stevens2024bioclip} which are adapted versions of CLIP that explicitly target natural world use cases.

\noindent{\bf Reranking with Multimodal Models.}
Reranking is a common paradigm in text retrieval, where a rapid search through pre-computed document indexes for potential matches is followed by a more expensive reranking of the top retrievals~\cite{nogueira2019passage,khattab2020colbert,karpukhin2020dense}.
In the image domain, reranking has been comparatively rare as the types of datasets for which it can be used are limited. 
In our experiments, we show that multimodal language models such as LLaVA~\cite{liu2024llavanext}, VILA~\cite{lin2023vila}, and GPT-4~\cite{achiam2023gpt,OpenAI2024} are effective rerankers out-of-the-box. 
To adapt these multimodal models for ranking, which requires a continuous score for a given text query and image pair, we prompt: \textit{Does this image show \{some query\}? Answer with ``Yes" or ``No" and nothing else.} (precise prompting details used for each model can be found in Appendix~\ref{appendix-lmm-prompting}). 
The logits of the ``Yes" and ``No" tokens are then used to compute the score:  
$s = s_y / (s_y + s_n)$, where  $s_y = \exp(logit_{Yes})$ and $s_n = \exp(logit_{No})$.

\section{Results} \label{section-results}
\vspace{-5pt}

Here we present a comprehensive evaluation of retrieval methods on \inquire{}. All results are reported on the test set. Additional results, including on the validation set, are in Appendix ~\ref{appendix-per-category-results}.%

\subsection{Metrics}

We evaluate using Average Precision at k (AP@k), Normalized Discounted Cumulative Gain (nDCG), and Mean Reciprocal Rank (MRR).  We primarily discuss AP as we find that this metric is the most discriminative of model performance. 
While these metrics have been commonly used to evaluate text retrieval, especially in the context of large-scale document retrieval~\cite{voorhees2005trec,craswell2020overview}, they have not found use in image retrieval due to the nonexistence of benchmarks like \inquire{} containing many relevant images for retrieval, rather than just one. 
Thus, we include them in our analysis to encourage their use in future image retrieval research. We note that the utilized AP@k metric uses a modified normalization factor suited to the retrieval setting.

Existing image retrieval benchmarks typically evaluate using the recall@k metric (\eg~\cite{li2023blip}), measuring if any of the top k images are relevant. 
While this makes sense in the setting where just one image is relevant, \inquire{} has potentially many relevant images and thus, we employ metrics that measure both relevance and ranking of retrievals.
Detailed discussion our metrics is provided in Appendix~\ref{appendix-metrics}.

\definecolor{lightergray}{rgb}{0.94, 0.94, 0.94}
\definecolor{eddieblue}{rgb}{0.75, 0.93, 0.99}

\begin{table}
\centering
\caption{Results for the \textsc{Inquire-Fullrank} task using two-stage retrieval. The top-k images are retrieved with CLIP ViT-H/14 and then reranked with the selected large multimodal models. Reranking offers a significant avenue of improvement.
}
\vspace{0.5em}
\small
\setlength{\tabcolsep}{0.5em}
\renewcommand{\arraystretch}{1.1}
\begin{tabular}{Slcccccc}
\toprule
 & \multicolumn{3}{c}{\textbf{Rerank Top 50}} & \multicolumn{3}{c}{\textbf{Rerank Top 100}} \\
 \cmidrule(l{3pt}r{3pt}){2-4} \cmidrule(l{3pt}r{3pt}){5-7}
\textbf{Method} & \textbf{mAP@50} & \textbf{nDCG@50} & \textbf{MRR} & \textbf{mAP@50} & \textbf{nDCG@50} & \textbf{MRR} \\ 
\rowcolor{lightergray}
\hline \textit{Initial ranking (ViT-H/14)} & 33.3 & 48.8 & 0.69 & 33.3 & 48.8 & 0.69 \\
\rowcolor{lightergray}
\textit{Best possible rerank} & 50.3 & 60.2 & 0.94 & 65.6 & 72.7 & 0.96 \\ \hline \noalign{\vskip 3pt} 
\multicolumn{7}{l}{\textit{Open-source multimodal models}} \\
BLIP-2 FLAN-T5-XXL~\cite{li2023blip} & 32.7 & 47.7 & 0.62 & 31.2 & 46.5 & 0.58 \\
InstructBLIP-T5-XXL~\cite{instructblip} & 34.1 & 49.0 & 0.67 & 33.0 & 48.3 & 0.64 \\
PaliGemma-3B-mix-448~\cite{beyer2024paligemma} & 35.0 & 49.7 & 0.70 & 35.6 & 50.6 & 0.68 \\
LLaVA-1.5-13B~\cite{liu2023improved} & 33.1 & 48.4 & 0.66 & 32.2 & 47.9 & 0.64 \\
LLaVA-v1.6-7B~\cite{liu2024llavanext} & 33.3 & 48.4 & 0.66 & 32.3 & 47.9 & 0.62 \\
LLaVA-v1.6-34B~\cite{liu2024llavanext} & 34.8 & 49.7 & 0.69 & 35.7 & 51.2 & 0.69 \\
VILA-13B~\cite{lin2023vila} & 35.0 & 49.6 & 0.67 & 35.7 & 50.8 & 0.65 \\
VILA-40B~\cite{lin2023vila} & \textbf{37.4} & \textbf{51.4} & \textbf{0.73} & \textbf{40.2} & \textbf{54.6} & \textbf{0.72} \\
\midrule \multicolumn{7}{l}{\textit{Proprietary multimodal models}} \\
\ \ GPT-4V~\cite{achiam2023gpt} & 35.8 & 50.7 & 0.73 & 36.5 & 51.9 & 0.72 \\
\ \ GPT-4o~\cite{OpenAI2024} & \textbf{39.6} & \textbf{53.4} & \textbf{0.79} & \textbf{43.7} & \textbf{57.9} & \textbf{0.78} \\
\bottomrule
\end{tabular}
\label{tab:rerank_results}
\vspace{-8pt}
\end{table}

\subsection{Fullrank Retrieval Task Results}
\vspace{-5pt}
We report full retrieval evaluation on \inquire{} in Tables~\ref{tab:zero_shot_results} and \ref{tab:rerank_results}. 
The per-category performance of selected CLIP models is reported in Figures~\ref{fig:by_category_results} and~\ref{fig:by_supercategory_results}. Further detailed results are in Appendix~\ref{appendix-per-category-results}.

\noindent{\bf The best CLIP models leave significant room for improvement.} 
Table~\ref{tab:zero_shot_results} shows that the top performing CLIP model achieves a moderate mAP@50 of 35.6.  
Although scaling models increases performance (see Figure~\ref{fig:by_category_results}), these results suggest that just scaling might not be enough, so future research should seek methods to better incorporate domain knowledge. %

\noindent{\bf Small models struggle to answer many queries.} 
In Table~\ref{tab:zero_shot_results} we can see that CLIP RN50 and CLIP ViT-B-32 score an mAP@50 of just 7.6 and 8.2 respectively, demonstrating that these smaller models are unable to provide accurate retrievals for nearly all queries. Since the largest models get comparatively much higher scores, the queries are not impossible but rather difficult for smaller models.
DFN ViT-B-16, trained with curated data, outperforms the larger OpenAI ViT-L-14, emphasizing the opportunity to improve the performance of efficient models via better data or training methods.

\begin{figure}
\begin{subfigure}[h]{0.47\linewidth}
\includegraphics[width=\linewidth]{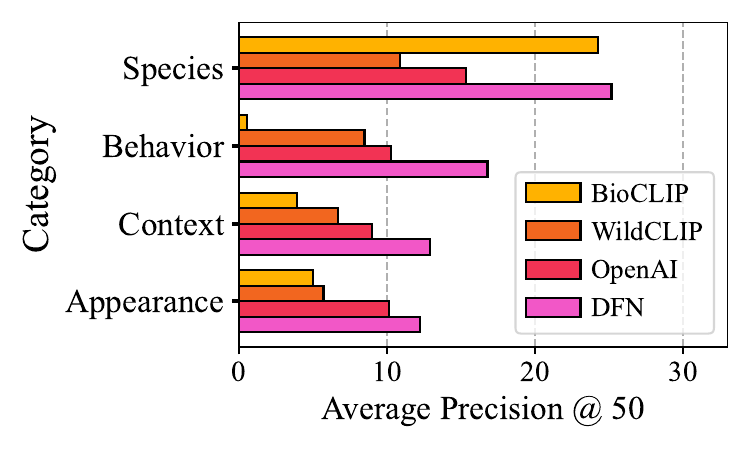}
\end{subfigure}
\hfill
\begin{subfigure}[h]{0.47\linewidth}
\includegraphics[width=\linewidth]{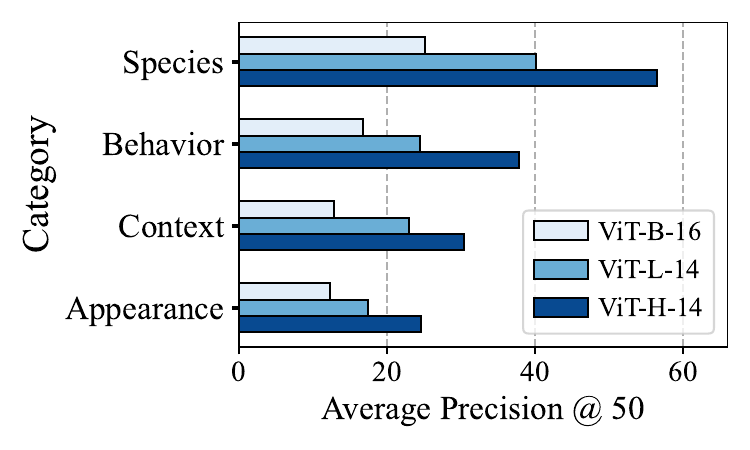}
\end{subfigure}%
\vspace{-0pt}
\caption{\textbf{Left:} CLIP zero-shot retrieval performance across supercategories using an identical backbone (ViT-B/16) trained or fine-tuned on different datasets. We see how training datasets have a significant effect on final performance, \eg BioCLIP is tuned on natural world data at the expense of forgetting other categories. \textbf{Right:} CLIP retrieval performance of models trained on DFN~\cite{fang2023data}.}
\label{fig:by_category_results}
\vspace{-6pt}
\end{figure}

\begin{figure}
\centering
\includegraphics[width=0.95\linewidth]{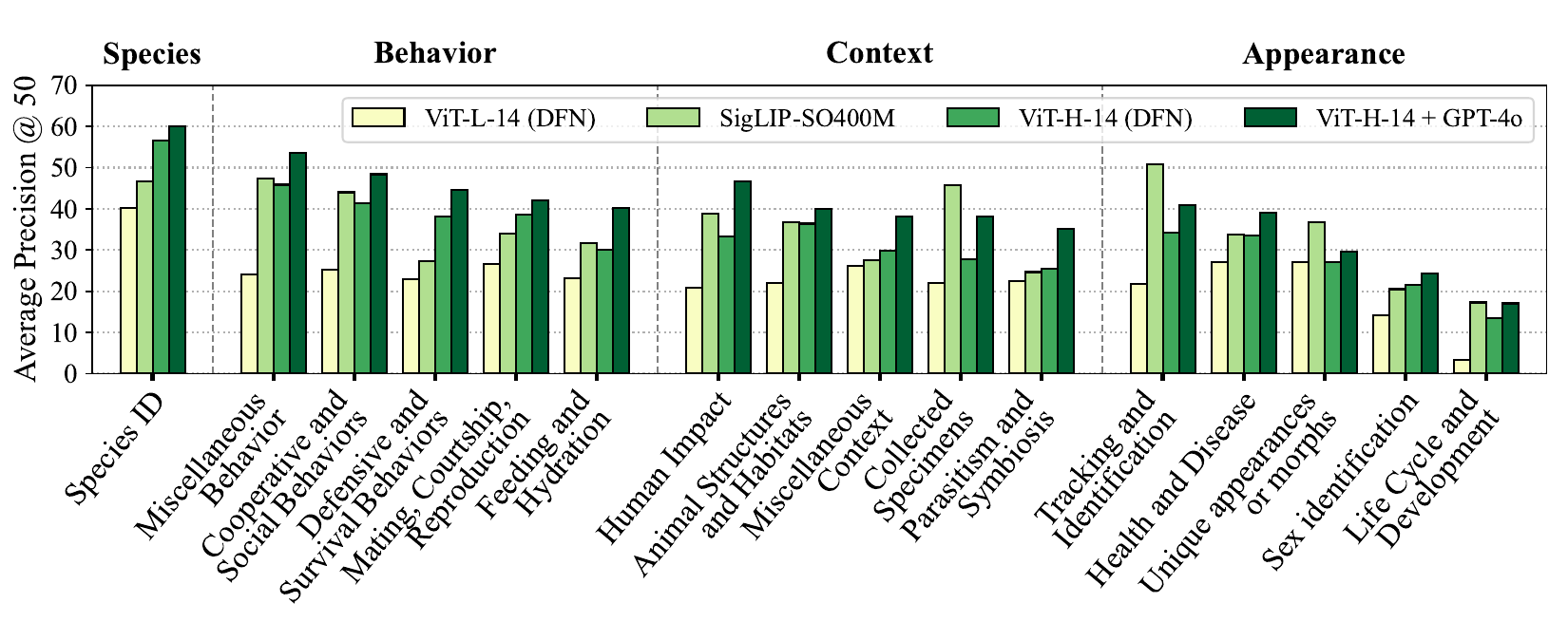}
\vspace{-6pt}
\caption{Retrieval performance for selected methods on \inquire{} query categories, ordered by difficulty. Some categories like \textsc{Life Cycle and Development} are exceptionally hard for current models. GPT-4o reranking improves performance in every category over its initial ViT-H-14 ranking.}
\label{fig:by_supercategory_results}
\vspace{-6pt}
\end{figure}

\noindent{\bf High-quality training data is crucial for expert-level queries}.
In Figure~\ref{fig:by_category_results}-left we show the retrieval performance on different supercategories for CLIP ViT-B/16 models that are trained on different datasets: BioCLIP~\cite{stevens2024bioclip}, WildCLIP~\cite{wildCLIP2024}, OpenAI~\cite{radford2021learning}, and DFN~\cite{fang2023data}. 
The DFN model, trained on two billion filtered image-text pairs, is the best generalist model on OpenCLIP's benchmarks~\cite{ilharco_gabriel_2021_5143773} and also outperforms all the others here, demonstrating the effectiveness of high quality pretraining. 
Conversely, models specifically trained on natural world data demonstrate degraded performance: BioCLIP was trained primarily on taxonomic captions and images, including iNat21, yet fails significantly on non-species queries, while WildCLIP has degraded performance in all supercategories.
This performance emphasizes the need for better natural world models and fine-tuning strategies that can gain domain-specific expertise while preserving generalist capabilities.

\noindent{\bf Reranking offers a valuable opportunity for improving retrieval.} Table~\ref{tab:rerank_results} shows that reranking with larger models like VILA-40B and GPT-4o gives a significant performance boost in mAP@50 of 7 and 12 points, respectively. Still, even GPT-4o performs significantly worse than the best possible rerank of its initial CLIP ViT-H-14 ranking. Increasing the size of the initial retrieval set from 50 to 100 can further improve performance by surfacing more relevant images, but only higher-performing models benefit: The mAP@50 for GPT-4o  increases by 5 points, while lower-performing models like LLaVA-v1.6-7B see decreased performance. Further results for varying the initial ranking set size are in Appendix~\ref{appendix-per-category-results}. Figure~\ref{fig:by_supercategory_results} visualizes how GPT-4o reranking improves performance on every category compared to its initial ViT-H-14 ranking.

\noindent{\bf Different query types present challenges of varying difficulties to existing models.} Figure~\ref{fig:by_supercategory_results} illustrates the difference in performance across query categories. 
We see that \textsc{Appearance} queries, which often require both domain knowledge of an organism's appearance and the fine-grained visual reasoning to recognize them, are the most difficult for existing models. 
Indeed, the \textsc{Life Cycle and Development} set (\eg \textit{``Immature bald eagle"}, \textit{``A cicada in the process of shedding its exoskeleton"}) are by far the most difficult. Conversely, \textsc{Context} queries such those in the \textsc{Human Impact} set (\eg \textit{``leopard on a road"}, \textit{``bird caught in a net"}), for which less expertise and comparatively coarser image understanding are needed, are easier for existing models.

\definecolor{lightergray}{rgb}{0.94, 0.94, 0.94}
\definecolor{eddieblue}{rgb}{0.75, 0.93, 0.99}

\begin{table}
\centering
\caption{Results for the \textsc{Inquire-Rerank} task on various embedding and multimodal models. For each task, a fixed set of the top-100 images is provided, which we then rerank using different methods. Evaluation metrics are calculated based solely on this fixed set, disregarding any potential positives outside of the top-100 images. Therefore, a perfect score is achievable within this context.}
\vspace{0.5em}
\small
\setlength{\tabcolsep}{0.4em}
\renewcommand{\arraystretch}{1.1}

\begin{minipage}{.47\linewidth}
\centering

\begin{tabular}{Sllccc}
\toprule
\textbf{Method} & \textbf{AP} & \textbf{nDCG} & \textbf{MRR} \\ \hline
\rowcolor{lightergray}
\textit{Random} & 22.1 & 52.6 & 0.35 \\ \hline
\noalign{\vskip 3pt} 
\multicolumn{4}{l}{\textit{Embedding models}} \\

\ \ CLIP ViT-B-32~\cite{radford2021learning}          & 30.2 & 59.1 & 0.47 \\
\ \ CLIP ViT-L-14~\cite{radford2021learning}          & 36.8 & 64.2 & 0.57 \\
\ \ CLIP ViT-H-14~\cite{fang2023data}             & 42.6 & 68.7 & 0.66 \\
\ \ SigLIP SO400m-14~\cite{zhai2023sigmoid}       & \textbf{50.1} & \textbf{73.5} & \textbf{0.72} \\

\midrule
\multicolumn{4}{l}{\textit{Proprietary multimodal models}} \\

\ \ GPT-4V~\cite{achiam2023gpt}   &  47.8 & 71.9 & 0.70 \\
\ \ GPT-4o~\cite{OpenAI2024}   &  \textbf{59.6} & \textbf{78.9} & \textbf{0.78} \\
\bottomrule
\end{tabular}

\end{minipage}%
\hfill
\begin{minipage}{.53\linewidth}
\centering
\begin{tabular}{Sllccc}
\toprule
\textbf{Method} & \textbf{AP} & \textbf{nDCG} & \textbf{MRR} \\  \hline \noalign{\vskip 5.55pt} 
\multicolumn{4}{l}{\textit{Open-source multimodal models}} \\ 

\ \ BLIP-2 T5-XXL~\cite{li2023blip}                   & 40.0 & 65.4 & 0.55 \\
\ \ InstructBLIP-T5-XXL~\cite{instructblip}    & 41.5 & 66.9 & 0.59 \\
\ \ PaliGemma-3B-mix-448~\cite{beyer2024paligemma}         & 42.9 & 67.9 & 0.60 \\
\ \ LLaVA-v1.5-13B~\cite{liu2023improved}             & 43.7 & 68.4 & 0.61 \\
\ \ LLaVA-v1.6-7B~\cite{liu2024llavanext}             & 46.9 & 70.4 & 0.66 \\
\ \ LLaVA-v1.6-34B~\cite{liu2024llavanext}            & 47.0 & 70.4 & 0.62 \\
\ \ VILA-13B~\cite{lin2023vila}                       & 47.1 & 71.1 & 0.67 \\
\ \ VILA-40B~\cite{lin2023vila}                       & \textbf{52.8} & \textbf{74.4} & \textbf{0.71} \\

\noalign{\vskip 3pt} 
\bottomrule
\end{tabular}
\end{minipage} 
\label{tab:rerank_challenge_results}
\vspace{-6pt}
\end{table}

\subsection{Rerank Retrieval Task Results}
\vspace{-4pt}
The results for the \textsc{INQUIRE-Rerank} task are presented in Table~\ref{tab:rerank_challenge_results}, where we evaluate reranking performance of both CLIP-style models like ViT-B-32 and larger vision-language models such as GPT-4o. Since the total number of images for each query is small (\ie 100), we also show the expected results of a random reranking for baseline comparison. In Table~\ref{tab:rerank_by_category} we further break down \textsc{INQUIRE-Rerank} results by queries containing scientific terminology and by query supercategory. 

\textbf{Current models struggle with expert-level text-to-image retrieval on \inquire{}.} In Table~\ref{tab:rerank_challenge_results} we observe that the highest AP of 59.6, achieved by GPT-4o, is far below the perfect score of 100, showing substantial room for improvement. Smaller models like CLIP ViT-B-32 only slightly outperform random chance. Since the top retrieved images are often visually or semantically similar, lower-performing models may be confused into promoting irrelevant images, leading to poorer ranking.

\textbf{Queries with scientific terminology are significantly more challenging, showing that models might not understand domains-specific language}. For example, the query\textit{``Axanthism in a green frog''}---referring to a mutation limiting yellow pigment production, resulting in a blue appearance---uses specialized terminology that a model may not understand. As a result, a model may incorrectly rank typical green frogs higher than axanthic green frogs, leading to worse-than-random performance. We show the performance of reranking models on queries with scientific terminology in Table~\ref{tab:rerank_by_category}. Interestingly, GPT-4o appears to be closing this gap, with an average difference of 7 points between queries with and without scientific terminology (AP scores of 53 and 60, respectively), compared to a 16-point difference for the next best model, VILA-40B (AP of 39 and 55). Nevertheless, this gap remains. Future work should explore methods to improve models' comprehension of domain-specific language, which is critical for accurate retrieval in scientific contexts.

\textbf{Reranking effectiveness varies widely by the query type.} Table~\ref{tab:rerank_by_category} shows that \textsc{Context} queries, often requiring general visual understanding, benefit substantially from reranking. Conversely, \textsc{Species} queries, requiring fine-grained visual understanding, see minimal improvement, with the specialized BioCLIP beating out even GPT-4o. These trends suggest that while recent models have better generalized vision capabilities, they continue to struggle with fine-grained visual understanding.

\begin{table}[]
    \centering
    \caption{Evaluation of \textsc{INQUIRE-Rerank} with queries grouped into different query types. First, we group queries containing scientific lingo and no scientific lingo. Next, we group queries by their supercategory (Appearance, Behavior, Context, Species). Queries with lingo tend to be more difficult, especially for large models with good generalist understanding but lacking domain expertise. All results are reported in AP.}
    \vspace{0.5em}
    \label{tab:rerank_by_category}
    \small
    \renewcommand{\arraystretch}{1.1}
    \begin{tabular}{lcccccc}
        \toprule
         & \multicolumn{2}{c}{\textbf{By Lingo}} & \multicolumn{4}{c}{\textbf{By Supercategory}}  \\  
    \cmidrule(r){2-3} 
    \cmidrule(r){4-7}
    Model & Lingo & No Lingo & Appearance & Behavior & Context & Species \\ \midrule
WildCLIP                              & 20.8 & 32.0 & 31.8 & 29.8 & 35.7 & 36.0 \\
BioCLIP                                  & 17.2 & 30.3 & 27.8 & 25.6 & 31.3 & \textbf{44.8} \\
CLIP ViT-B-32                                 & 22.8 & 31.6 & 29.5 & 31.0 & 36.0 & 35.3 \\
CLIP ViT-L-14                                 & 29.2 & 37.4 & 37.2 & 36.2 & 40.2 & 38.2 \\
CLIP ViT-H-14                             & 32.1 & 44.0 & 38.1 & 50.5 & 45.5 & 31.2 \\
SigLIP SO400m-14                     & \textbf{38.3} & \textbf{51.7} & \textbf{51.7} & \textbf{53.6} & \textbf{54.1} & 44.6 \\
\midrule
LLaVA-v1.6-34B               & 28.2 & 49.5 & 41.0 & 48.2 & 53.8 & \textbf{37.5} \\
VILA-13B        & 37.2 & 48.2 & 39.3 & 47.1 & 58.6 & 34.8 \\
VILA-40B        & \textbf{38.6} & \textbf{54.5} & \textbf{46.9} & \textbf{54.9} & \textbf{63.1} & 37.0 \\
\midrule
GPT-4V                 & 35.9 & 49.3 & 40.3 & 50.2 & 54.2 & 39.0 \\
GPT-4o                     & \textbf{53.3} & \textbf{60.4} & \textbf{51.9} & \textbf{61.4} & \textbf{75.4} & \textbf{44.3} \\
\bottomrule
    \end{tabular}
\end{table}

\section{Limitations and Societal Impact}\label{section-limitations}
\vspace{-5pt}
While the species labels for each image in iNat24 are generated via consensus from multiple citizen scientists, there may still be errors in the labels which our evaluation will inherit. 
However, this error rate is estimated to be low~\cite{loarie_2024}. 
\inquire{} contains natural world images, which while diverse, may hinder the relevance of some of our insights to other visual domains. 
In spite of this, we believe that due to the wide range of visual queries contained within, progress on \inquire{} will likely be indicative of multimodal model performance on other challenging domains. 

There could be unintended negative consequences if conservation assessments were made based on the predictions from biased or inaccurate models evaluated in this paper. 
Where relevant, we have attempted to flag these performance deficiencies. 
While we have filtered out personally identifiable information from our images, the retrieval paradigm allows for free-form text search and thus care should be taken to ensure that appropriate text filters are in-place to prevent inaccurate or hurtful associations being made between user queries and images of wildlife.

\section{Conclusion} 
\vspace{-5pt}
We introduced \inquire{}, a challenging new text-to-image retrieval benchmark which consists of expert-level text queries that have been exhaustively annotated across a large pool of five million natural world images called iNat24. 
This benchmark aims to emulate real world  image retrieval and analysis problems faced by scientists working with these types of large-scale image collections. 
Our hope is that progress on \inquire{} will drive advancements in the real scientific utility of AI systems.
Our evaluation of existing methods reveals that \inquire{} poses a significant challenge even for the current largest state-of-the-art multimodal models, showing there is significant room for innovations to develop accurate retrieval systems for complex visual domains.

\clearpage
\begin{ack}
\vspace{-7pt}
We wish to thank the many iNaturalist participants for continuing to share their data and also the numerous individuals who provided suggestions for search queries. 
Special thanks to Kayleigh Neil, Beñat Yañez Iturbe-Ormaeche, Filip Dorm, and Patricia Mrazek for data annotation. 
Funding for annotation was provided by the Generative AI Laboratory (GAIL) at the University of Edinburgh. 
EV and SB were supported in part by the Global Center on AI and Biodiversity Change (NSF OISE-2330423 and NSERC 585136). 
OMA was in part supported by a Royal Society Research Grant. 
OP and KJ were supported by the Biome Health Project funded by WWF-UK. 
\end{ack} 

\bibliography{main}

\clearpage
\appendix

\noindent{\bf\huge Appendix}

\definecolor{mycolor}{rgb}{0.122, 0.435, 0.698}

\newmdenv[innerlinewidth=0.5pt, roundcorner=4pt,linecolor=mycolor,innerleftmargin=6pt,
innerrightmargin=6pt,innertopmargin=6pt,innerbottommargin=6pt]{mybox}

\vspace{10pt}
\begin{mybox}
Please refer to the links below for the dataset, code repository, and website:

\begin{itemize}
    \item Website: \textcolor{blue}{\url{https://inquire-benchmark.github.io/}}
    \item GitHub: \textcolor{blue}{\url{https://github.com/inquire-benchmark/INQUIRE}}
    \item Data: \textcolor{blue}{\url{https://github.com/inquire-benchmark/INQUIRE/tree/main/data}}
\end{itemize}
\end{mybox}

\addtocontents{toc}{\protect\setcounter{tocdepth}{2}}
\setlength{\cftbeforesecskip}{6pt}
\tableofcontents

\setcounter{table}{0}
\renewcommand{\thetable}{A\arabic{table}}
\setcounter{figure}{0}
\renewcommand{\thefigure}{A\arabic{figure}}
\newcolumntype{R}[1]{>{\begin{turn}{90}\begin{minipage}{#1}\scriptsize}l%
<{\end{minipage}\end{turn}}%
}

\newpage

\section{\inquire{} Query Examples}
Below we include several queries from \inquire{} with their broad justification, small number of examples of relevant and not relevant images, and a detailed explanation of each image's relevance.

\begin{figure}[ht]
    \centering
    \includegraphics[width=\linewidth]{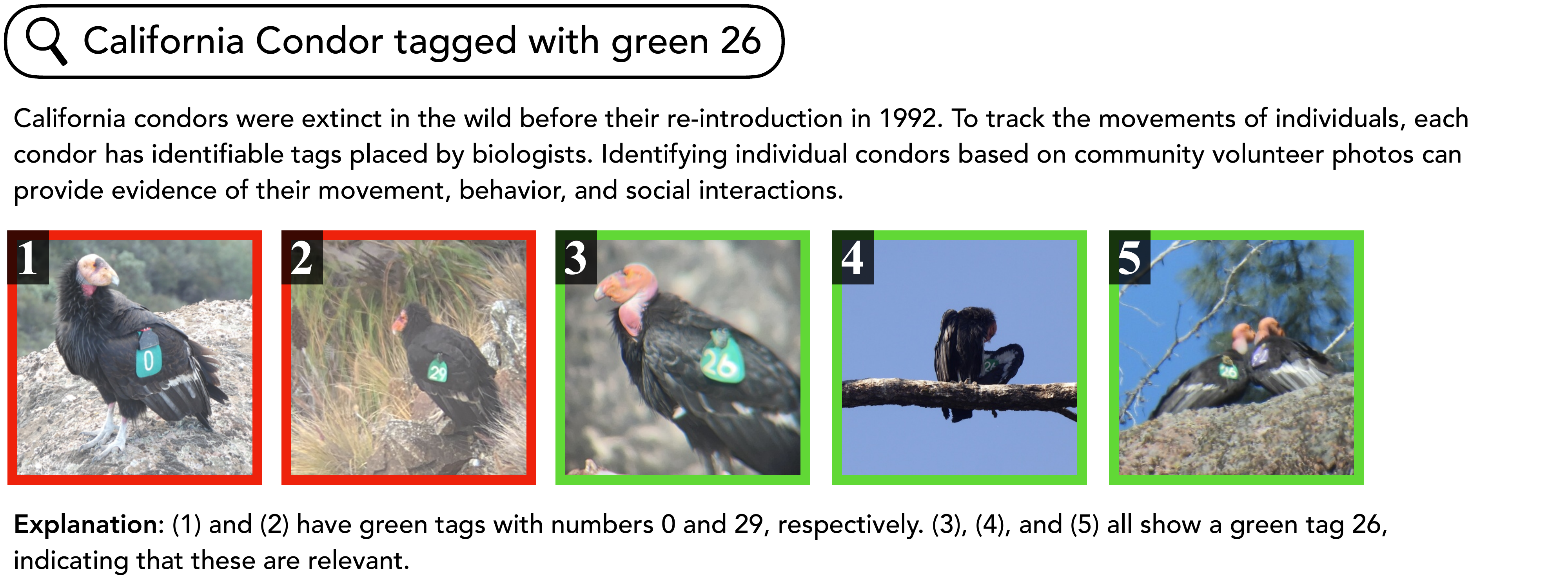} \\
    \noindent\rule{\textwidth}{1pt} \\
    \vspace{1.5em}
    \includegraphics[width=\linewidth]{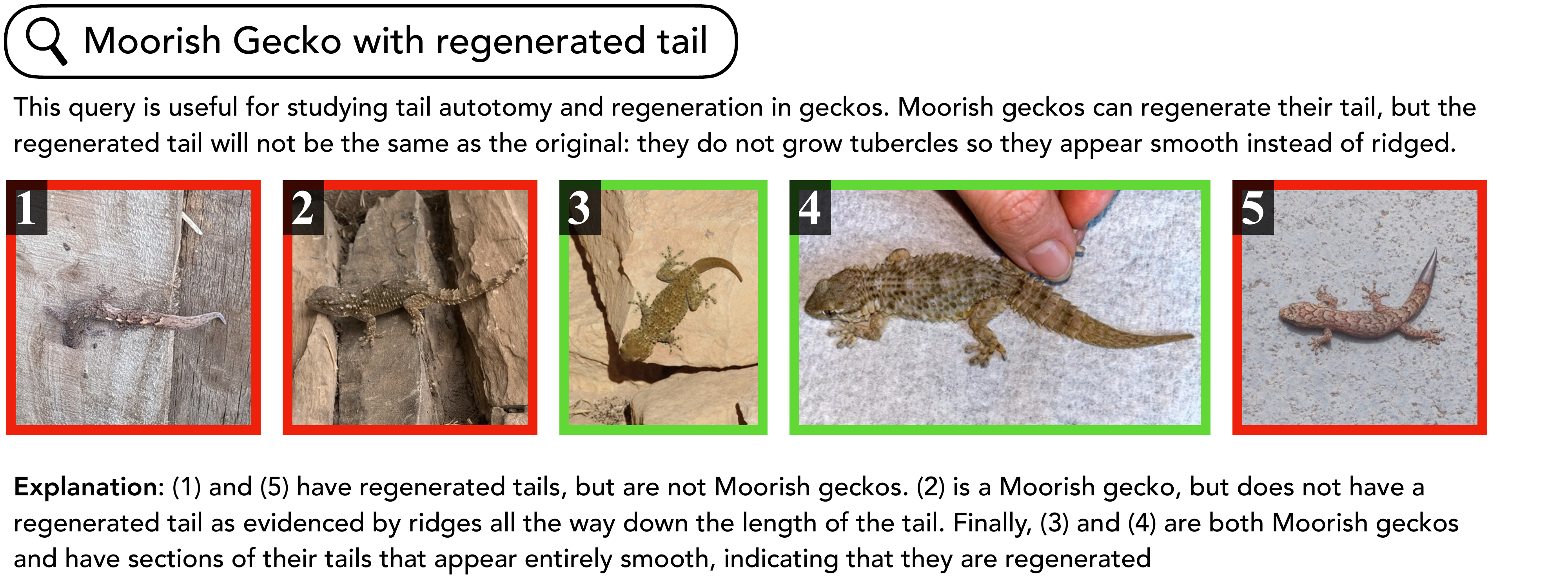} \\
    \noindent\rule{\textwidth}{1pt} \\
    \vspace{1.5em}
    \includegraphics[width=\linewidth]{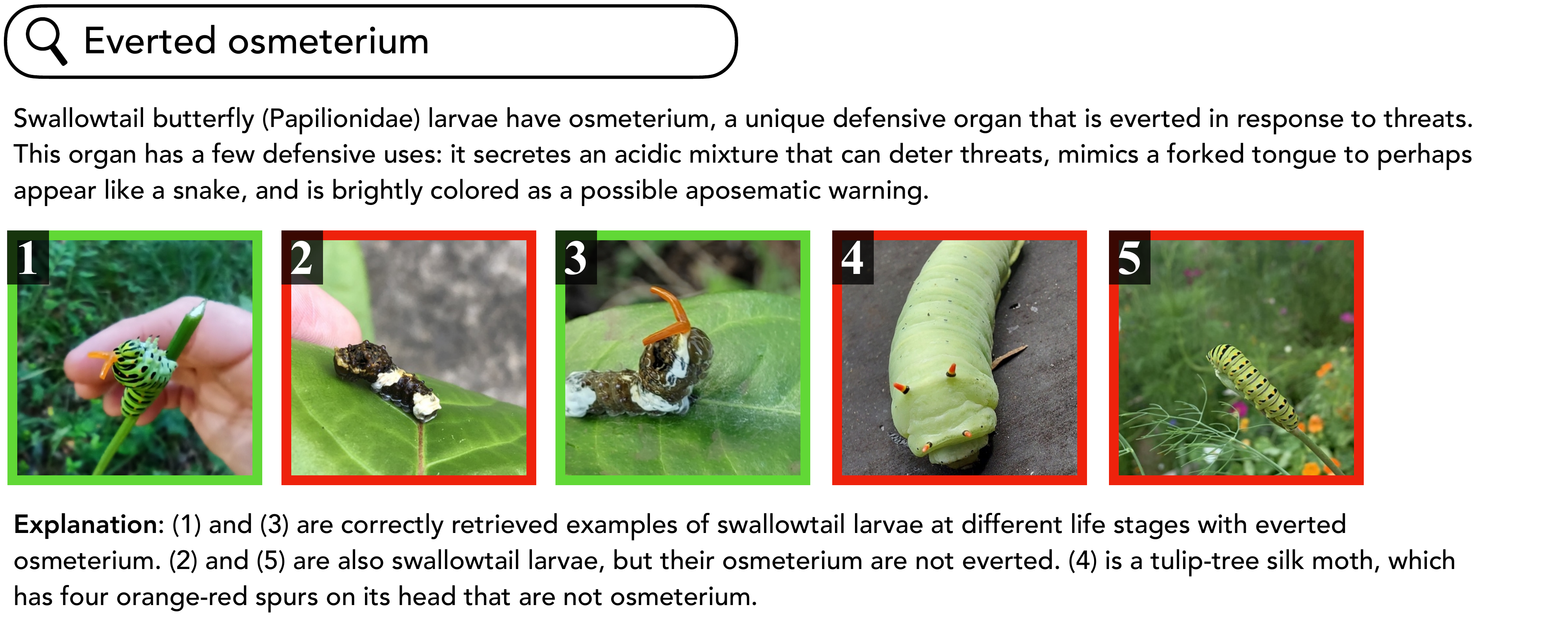}
    \label{fig:ex_queries_1}
\end{figure}

\begin{figure}
    \centering
    \includegraphics[width=\linewidth]{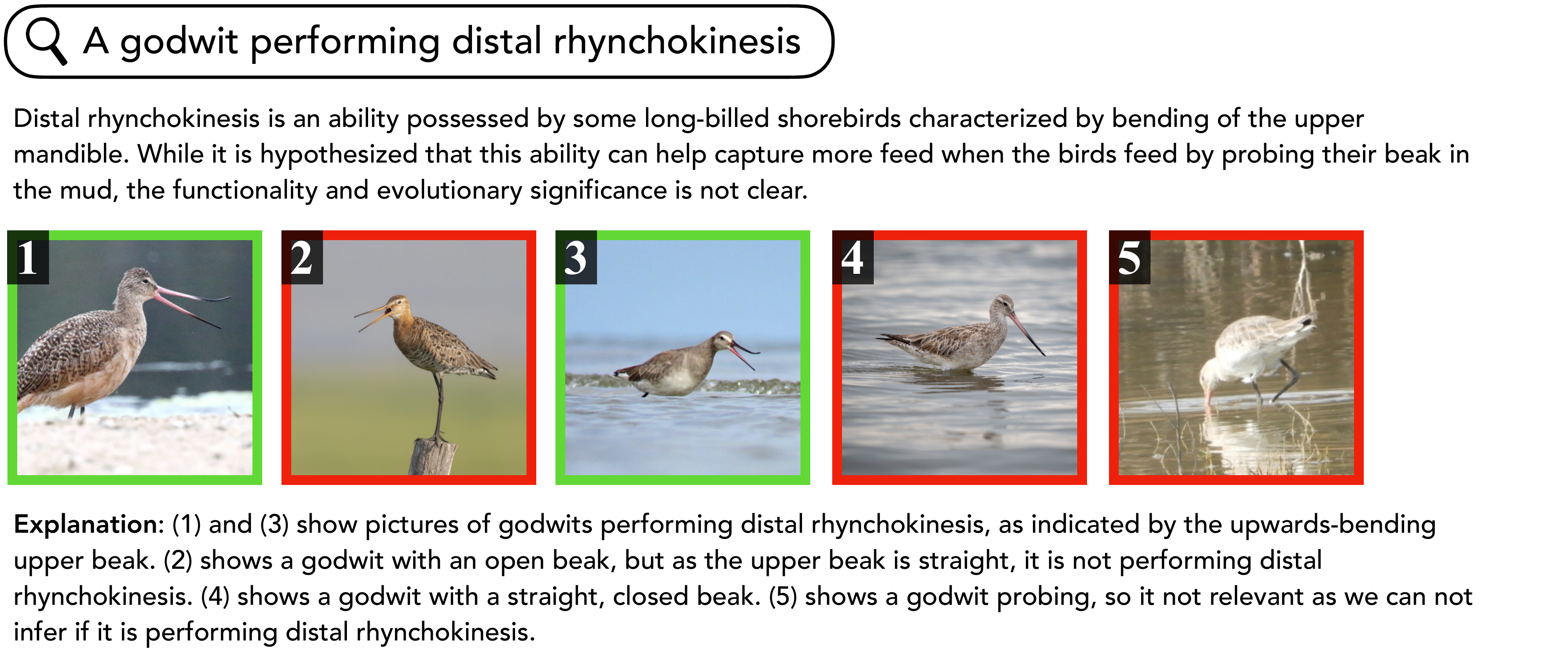} \\
    \noindent\rule{\textwidth}{1pt} \\
    \vspace{1.5em}
    \includegraphics[width=\linewidth]{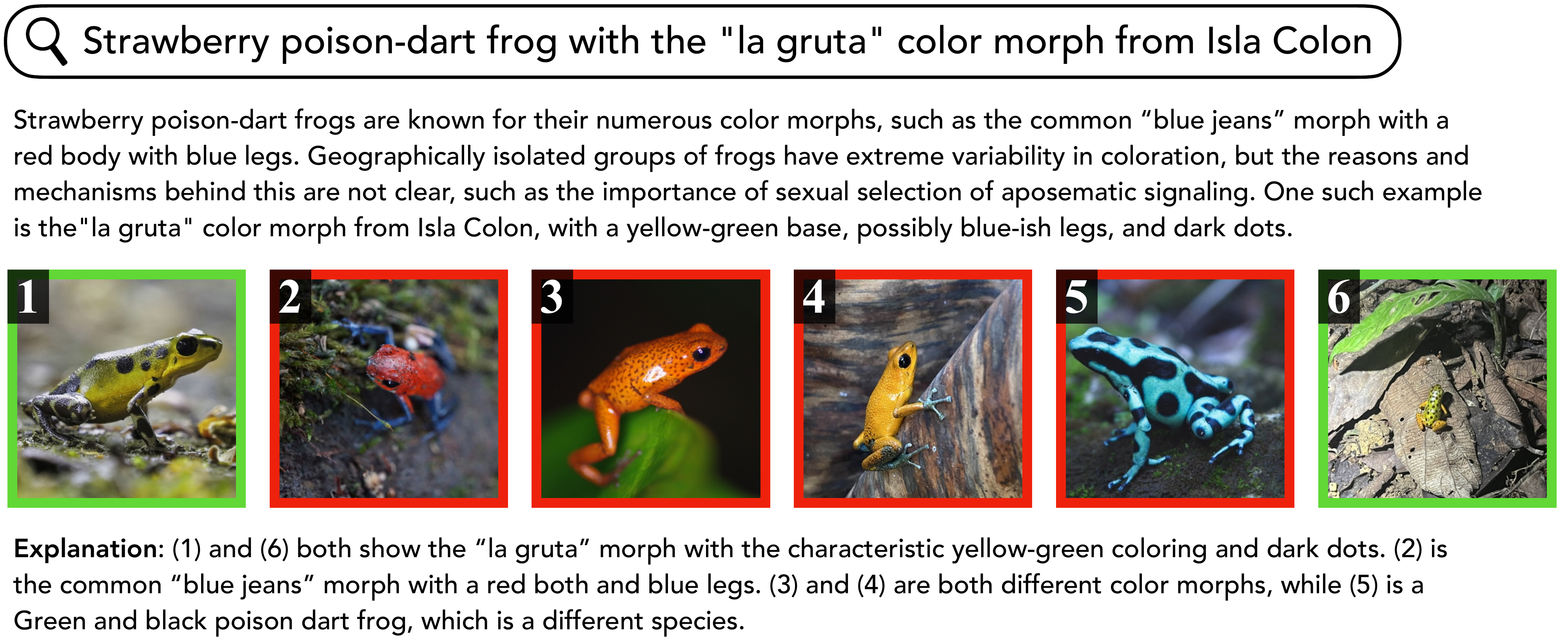} 
    \\
    \noindent\rule{\textwidth}{1pt} \\
    \vspace{1.5em}
    \includegraphics[width=\linewidth]{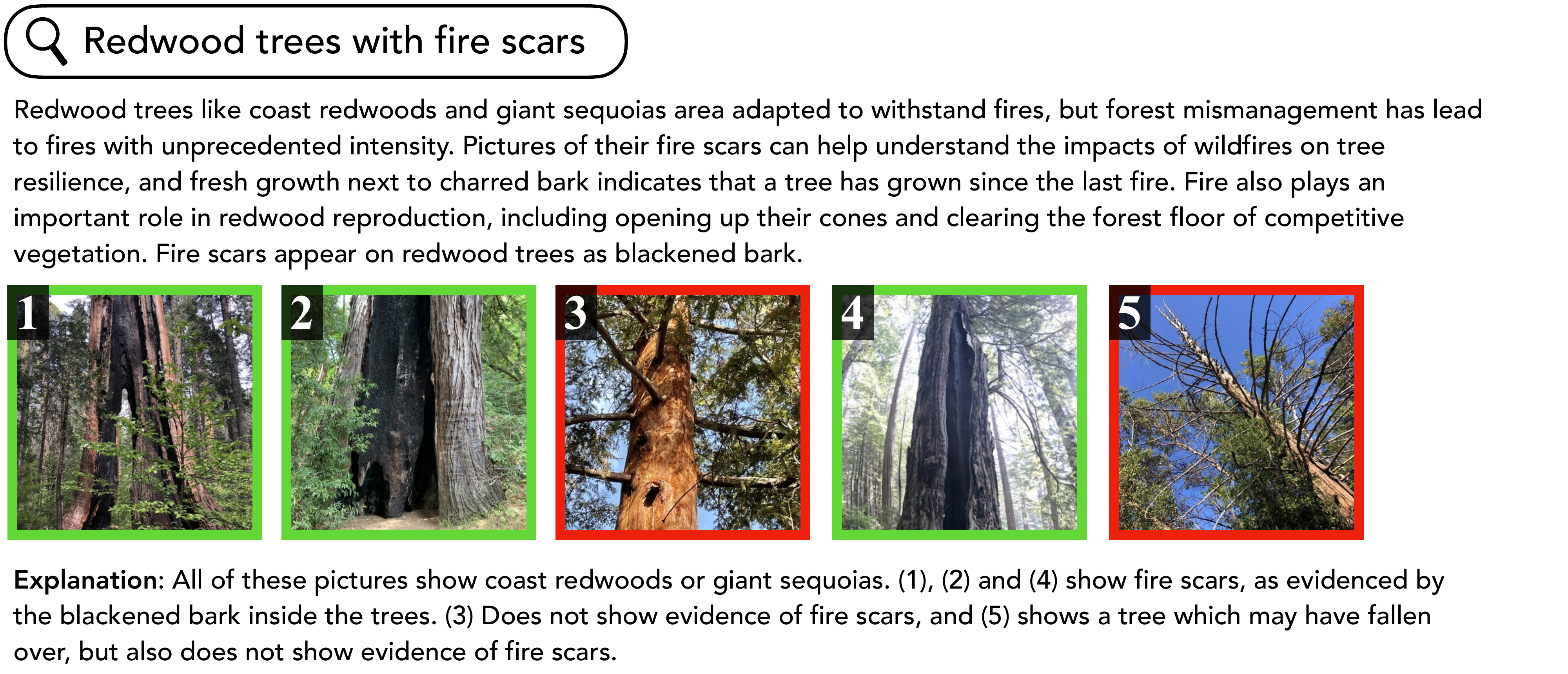} 
    \label{fig:ex_queries_2}
\end{figure}

\newpage
\section{Additional Details about \inquire{}}

\begin{figure}[ht]
\begin{subfigure}[h]{0.49\linewidth}
\includegraphics[width=\linewidth]{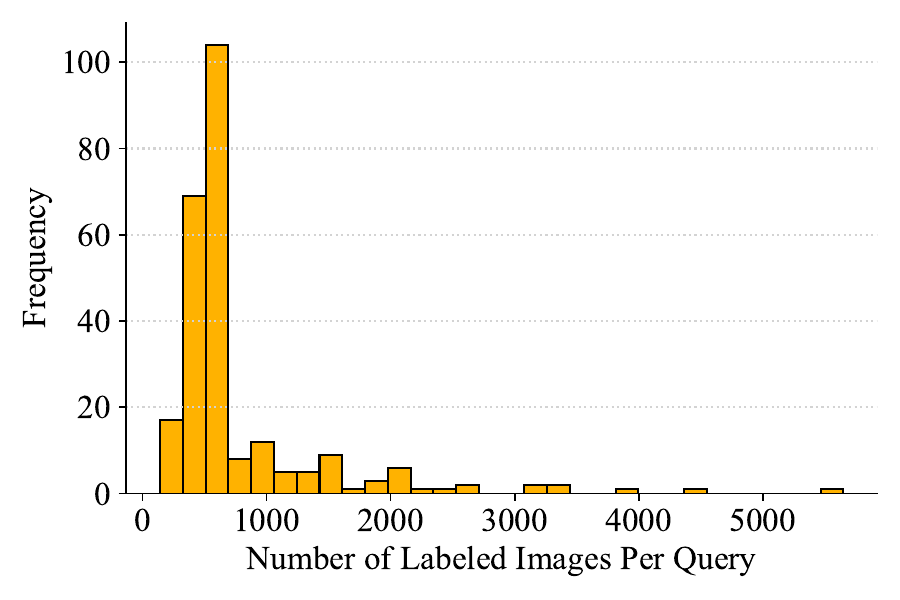}
\end{subfigure}
\hfill
\begin{subfigure}[h]{0.49\linewidth}
\includegraphics[width=\linewidth]{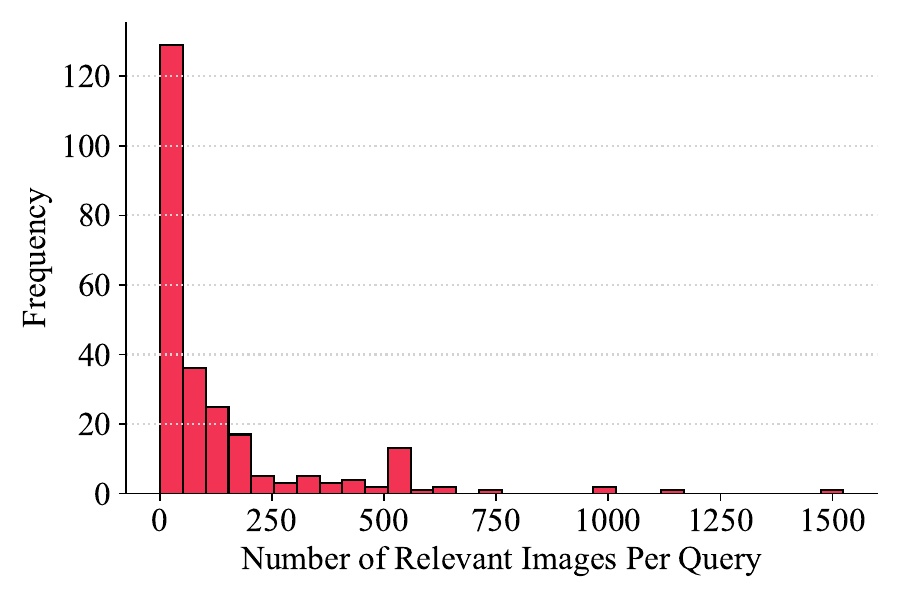}
\end{subfigure}%
\caption{Collecting \inquire{} involved labeling a combined total of 194,334 candidate images across all 250 queries, yielding 32,696 relevant matches. These histograms show the number of images labeled per query, and the number of those that were relevant. }
\label{fig:supp_histograms}
\end{figure}

In Figure~\ref{fig:supp_histograms} we show histograms representing the number of labeled images and relevant images for each query from \inquire{}. We see that there is a long-tailed distribution for the number of relevant images per query, which ranges from 1 to 1500, with an average of 131 and median of 46 relevant images per query. In total, we labeled 194,334 images, of which 32,696 were relevant to their queries (or 32,553 unique images). As we use species filters and other steps to ensure our labeling is comprehensive (see Appendix~\ref{appendix-data-annotation}), we treat the rest of the iNat24 images as not relevant. This means that along with the existing image labels, we also have about 5 million weak negative labels per query, for a total of 1 billion weak labels.

In Table~\ref{table:supp_all_categories} we provide a breakdown of the number of queries of each of the four main types, including the average number of relevant images for each query. We note that this number varies widely. Species queries tend to have many relevant labels, while queries in categories like "Tracking and Identification" tend to have few relevant images.

\begin{table}[ht]
    \centering
    \caption{\inquire{} queries can be grouped in to 16 categories. Here we provide a list of these categories, the number of queries in each, and an example.}
    \vspace{5pt}
      \resizebox{1.0\columnwidth}{!}{
  \renewcommand{\arraystretch}{1.2}
    \begin{tabular}{llb{1.4cm}b{1.4cm}b{1.4cm}l}
        \toprule
        \textbf{Supercategory} & \textbf{Category} & \shortstack[l]{\textbf{\# Val} \\ \textbf{Queries}} & \shortstack[l]{\textbf{\# Test} \\ \textbf{Queries}} & \shortstack[l]{\textbf{Mean \#} \\ \textbf{Relevant}} & \textbf{Example} \\ 
        \midrule
        \multirow{5}{*}{Appearance} & Health and Disease & 2 & 15 & 140 & red fox showing signs of sarcoptic mange  \\
                            & Life Cycle and Development & 4 & 18 & 104 & penguin during molting period \\
                            & Sex identification & 2 & 13 & 216 & fiddler crab with an oversized chela \\
                            & Tracking and Identification & 4 & 8 & 30 & California Condor tagged with green 26 \\
                            & Unique appearances or morphs & 5 & 12 & 79 & a male mandarin duck in breeding plumage \\ \midrule
\multirow{5}{*}{Behavior} & Cooperative and Social Behaviors & 4 & 20 & 49 & macaques engaging in mutual grooming behavior \\
                            & Defensive and Survival Behaviors & 2 & 11 & 51 & everted osmeterium \\
                            & Feeding and Hydration & 5 & 18 & 62 & A godwit performing distal rhynchokinesis \\
                            & Mating, Courtship, Reproduction & 4 & 11 & 39 & Alligator lizards mating \\
                            & Miscellaneous Behavior & 2 & 7 & 151 & spider monkey using its tail to hang on a branch \\ \midrule
\multirow{5}{*}{Context} & Animal Structures and Habitats & 4 & 8 & 230 & a rose bedeguar gall \\
                            & Collected Specimens & 1 & 5 & 34 & measuring the body dimensions of a bee \\
                            & Human Impact & 4 & 11 & 45 & dehorned rhin \\
                            & Miscellaneous Context & 2 & 13 & 139 & Mushrooms growing in a fairy ring formation \\
                            & Parasitism and Symbiosis & 1 & 12 & 56 & Sharks with remoras attached \\ \midrule
\multirow{1}{*}{Species} & Species ID & 4 & 18 & 520 & blue dragon nudibranch \\
        \bottomrule
    \end{tabular}
    }
    \label{table:supp_all_categories}
\end{table}

\begin{figure}[t]
    \centering
    \includegraphics[width=\linewidth]{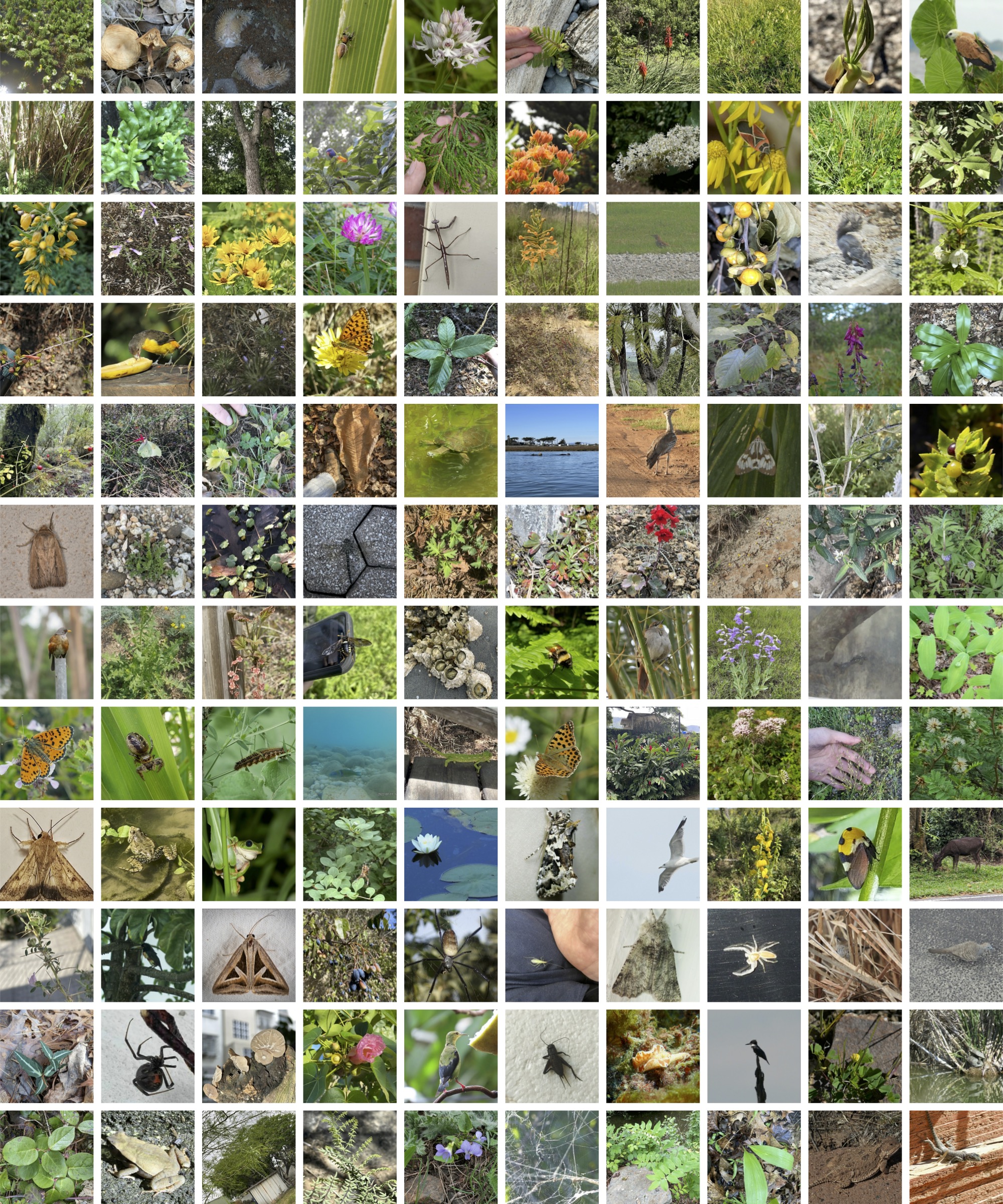}
    \caption{Random images from the iNat24 dataset. iNat24 contains five million images from 10,000 species classes.}
    \label{fig:inat24_teaser}
\end{figure}

\section{Additional Details about iNat24}

iNat24 contains 4,813,543 images for 9,959 species. Figure~\ref{fig:inat24_teaser} shows examples of randomly chosen images from the dataset.

\section{Geographic Range of \inquire{} and iNat24}
\begin{figure}
\centering
\begin{subfigure}[h]{\linewidth}
\includegraphics[width=\linewidth]{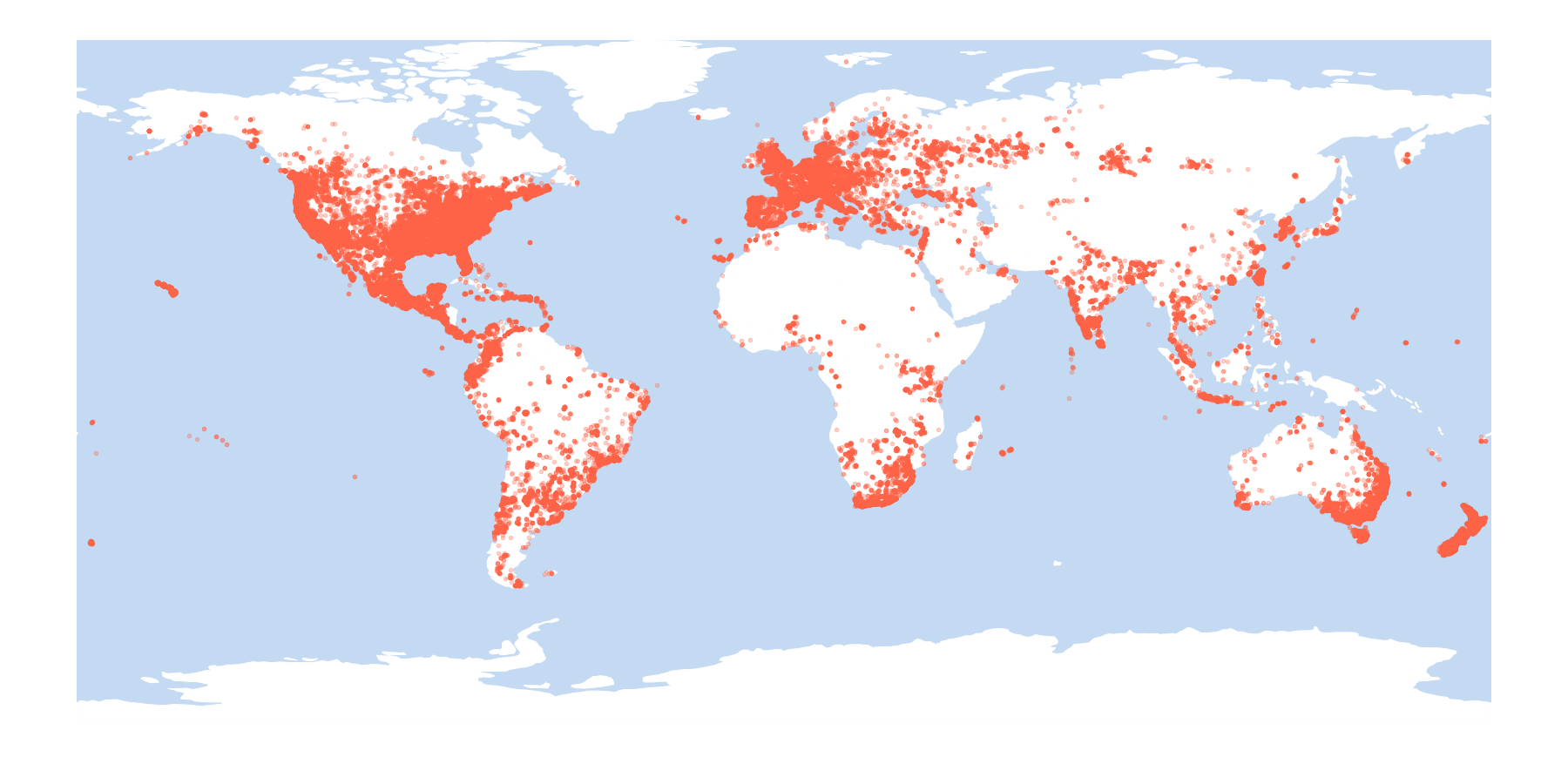}
\vspace{-2em}
\caption{Geographic distribution of all iNat24 images.}
\end{subfigure}
\hfill
\begin{subfigure}[h]{\linewidth}
\includegraphics[width=\linewidth]{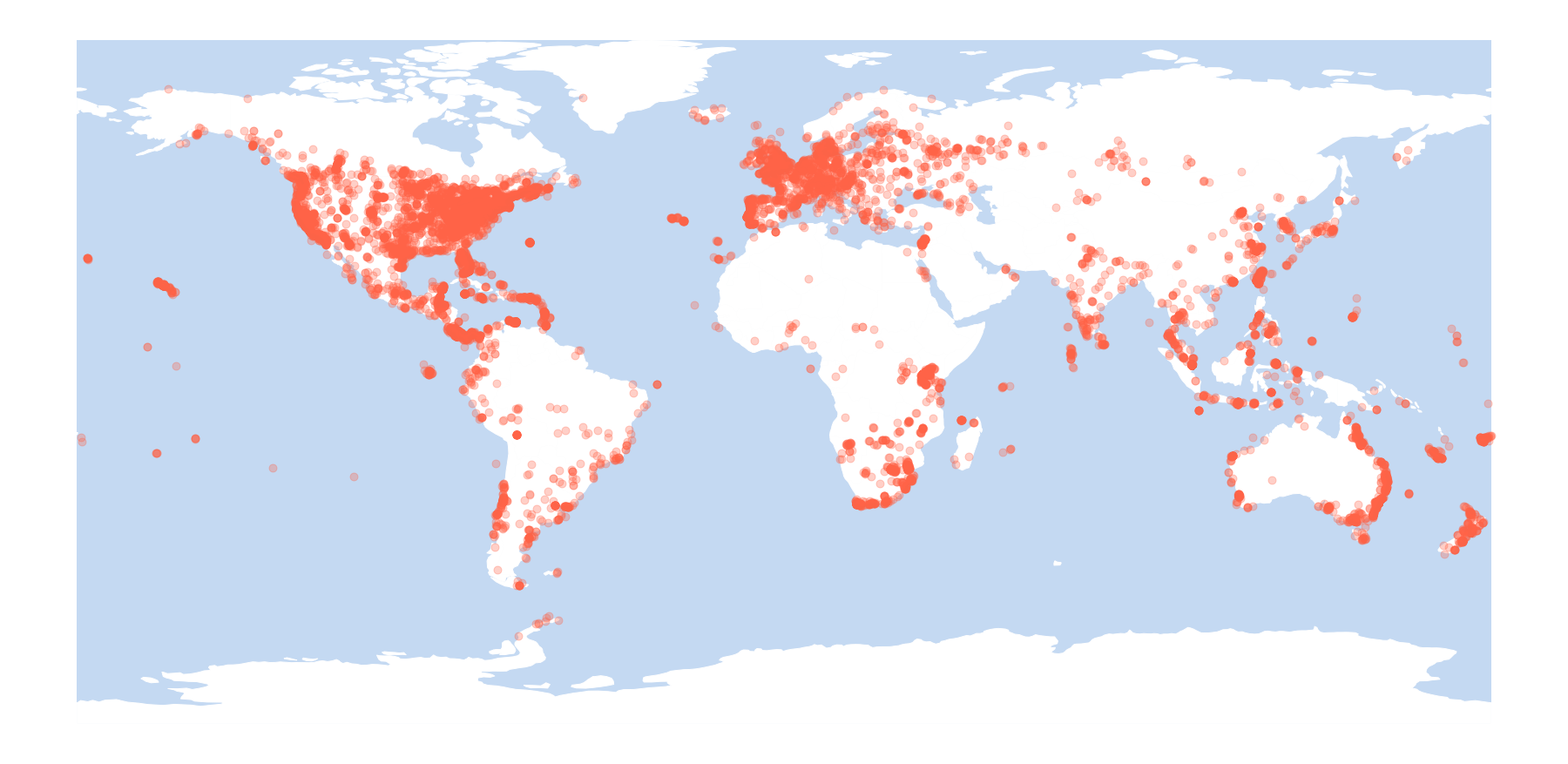}
\vspace{-2em}
\caption{Geographic distribution of the iNat24 images marked relevant for an \inquire{} query.}
\end{subfigure}%
\caption{Here we compare the spatial distribution of the images in iNat24 to the relevant images in queries from \inquire{}. 
We can see that the distribution of both is similar. 
Both exhibit a bias towards North America, Europe, and parts of Australasia which is reflective of the spatial biases present in the iNaturalist platform.}
\label{fig:appendix-map}
\end{figure}

In Figure~\ref{fig:appendix-map} we show the geographic range of iNat24 observations and image from \inquire{} judged as relevant. We can see that the distribution of both is similar, which demonstrates that \inquire{} queries do not exhibit a strong geographic bias as compared to the iNat24 source data in the images that the queries correspond to.
However, both exhibit a bias towards North America, Europe, and parts of Australasia which is reflective of the spatial biases present in the iNaturalist platform.

\section{Additional Results}\label{appendix-per-category-results}
In this section we provide detailed evaluation results on \textsc{INQUIRE-Fullrank} and \textsc{INQUIRE-Rerank}. 

\subsection{Breaking Down Results by Category}
All \inquire{} queries belong to one of sixteen fine-grained categories, so grouping results by each category provides insights into the specific strengths and weaknesses of current ranking methods. In Table~\ref{tab:results_fullrank_by_supercategory} we show the results on \textsc{INQUIRE-Fullrank} by category for all embeddings models we test. In Table~\ref{tab:results_rerank_by_supercategory} we show the results on\textsc{INQUIRE-Rerank} for all embeddings models and vision-language models we test. Both tables show results for test set queries.

\begin{table}[t]  
    \caption{Detailed evaluation of \textsc{INQUIRE-Fullrank} by category for a variety of embedding models. Results are reported in AP@50.}
    \label{tab:results_fullrank_by_supercategory}
    \vspace{4pt}

  \scriptsize
  \setlength\tabcolsep{3pt}
  \renewcommand{\arraystretch}{1.08}
  \centering
  \begin{tabular}{lcccccccccccccccc}  
    \toprule  
    Model & \multicolumn{1}{c}{Sp.}
     & \multicolumn{5}{c}{Behavior} &
      \multicolumn{5}{c}{Context} & \multicolumn{5}{c}{Appearance} \\  
    \cmidrule(r){2-2} 
    \cmidrule(r){3-7} \cmidrule(r){8-12} \cmidrule(r){13-17}
        & \multicolumn{1}{R{2cm}}{Species ID}
        & \multicolumn{1}{R{2cm}}{Miscellaneous \\Behavior}
        & \multicolumn{1}{R{2cm}}{Defensive and \\Survival Behaviors}
        & \multicolumn{1}{R{2cm}}{Cooperative and \\Social Behaviors}
        & \multicolumn{1}{R{2cm}}{Mating, Courtship, \\Reproduction}
        & \multicolumn{1}{R{2cm}}{Feeding and \\Hydration}
        & \multicolumn{1}{R{2cm}}{Human Impact}
        & \multicolumn{1}{R{2cm}}{Miscellaneous \\Context}
        & \multicolumn{1}{R{2cm}}{Animal Structures \\and Habitats}
        & \multicolumn{1}{R{2cm}}{Parasitism and \\Symbiosis}
        & \multicolumn{1}{R{2cm}}{Collected \\Specimens}
        & \multicolumn{1}{R{2cm}}{Tracking and \\Identification}
        & \multicolumn{1}{R{2cm}}{Health and \\Disease}
        & \multicolumn{1}{R{2cm}}{Unique appearances or morphs}
        & \multicolumn{1}{R{2cm}}{Sex identification}
        & \multicolumn{1}{R{2cm}}{Life Cycle and \\Development} \\
    \cmidrule(r){1-1} \cmidrule(rl){2-2} \cmidrule(rl){3-3} \cmidrule{4-4} \cmidrule(rl){5-5}
       \cmidrule(rl){6-6} \cmidrule{7-7} \cmidrule(rl){8-8} \cmidrule(rl){9-9} \cmidrule{10-10}
       \cmidrule(rl){11-11} \cmidrule{12-12} \cmidrule(rl){13-13} \cmidrule{14-14}
      \cmidrule(rl){15-15} \cmidrule(r{1pt}){16-16} \cmidrule(l{1pt}r{1pt}){17-17}
      bioclip                   & 24.2 &  2.8 &  0.0 &  0.8 &  0.1 &  0.0 &  0.0 &  7.7 & 10.4 &  0.8 &  0.0 &  1.0 & 10.1 &  5.7 &  6.6 &  1.0 \\
rn50                      & 10.2 & 13.8 &  5.4 &  6.6 &  3.0 &  6.2 &  2.7 & 10.3 &  7.3 &  4.8 &  6.5 &  2.2 & 12.5 & 14.0 &  2.7 &  1.7 \\
wildclip-t1t7-lwf         & 10.6 & 13.4 &  3.8 &  5.0 &  5.3 &  9.0 &  6.8 &  6.9 &  9.6 &  3.6 &  3.2 &  6.3 & 14.7 & 10.4 &  2.4 &  0.7 \\
wildclip-t1               & 10.9 & 14.4 &  5.7 &  8.7 &  4.3 & 10.2 &  5.8 &  7.6 & 11.3 &  4.0 &  5.7 &  5.5 & 12.6 &  6.8 &  3.3 &  1.1 \\
vit-b-32                  & 11.3 & 16.1 &  4.6 &  7.5 &  5.8 &  7.0 &  5.3 &  9.8 &  7.3 &  6.1 &  8.8 &  5.2 & 10.6 & 11.5 &  5.4 &  1.6 \\
vit-b-16                  & 15.3 & 14.8 &  8.3 & 12.0 &  7.6 &  9.4 & 10.4 &  7.9 & 11.3 &  5.8 & 12.8 & 10.5 & 18.1 & 17.2 &  6.4 &  1.4 \\
rn50x16                   & 18.1 & 21.9 & 15.0 & 15.5 &  6.7 & 16.4 & 11.6 & 13.8 & 11.0 &  8.1 & 14.2 & 14.5 & 22.5 & 16.0 &  6.3 &  5.7 \\
vit-l-14                  & 19.9 & 16.6 & 13.0 & 15.2 & 13.9 & 11.4 & 14.4 & 14.9 & 15.0 &  9.3 &  5.6 & 18.0 & 20.7 & 24.6 &  9.6 &  6.6 \\
vit-b-16-dfn              & 25.2 & 20.5 & 15.0 & 13.9 & 13.2 & 21.8 & 12.3 & 14.3 & 15.1 &  9.6 & 15.0 & 12.8 & 15.8 & 21.9 & 10.9 &  3.5 \\
vit-l-14-dfn              & 40.1 & 24.2 & 22.9 & 25.2 & 26.5 & 23.2 & 20.8 & 26.1 & 22.0 & 22.5 & 22.1 & 21.7 & 27.0 & 27.0 & 14.1 &  3.4 \\
siglip-vit-l-16-384       & 47.0 & 43.0 & 25.9 & 40.2 & 36.1 & 24.5 & 34.5 & \textbf{31.5} & 34.4 & 23.8 & 29.5 & 34.9 & \textbf{36.3} & 24.4 & 17.1 & \textbf{17.8} \\
vit-h-14-378              & \textbf{56.5} & 45.9 & \textbf{38.2} & 41.3 & \textbf{38.5} & 30.1 & 33.3 & 29.8 & 36.4 & \textbf{25.6} & 27.8 & 34.2 & 33.5 & 27.0 & \textbf{21.5} & 13.5 \\
siglip-so400m-14-384      & 46.6 & \textbf{47.4} & 27.4 & \textbf{44.0} & 33.9 & \textbf{31.7} & \textbf{38.8} & 27.6 & \textbf{36.8} & 24.7 & \textbf{45.6} & \textbf{50.9} & 33.8 & \textbf{36.7} & 20.5 & 17.3 \\
  \bottomrule
  \end{tabular}
  \vspace{-2em}
\end{table}

\begin{table}[t]  
    \caption{Detailed evaluation of \textsc{INQUIRE-Rerank} by category for a variety of embedding models and vision-language models. Results are reported in AP.}
    \label{tab:results_rerank_by_supercategory}
    \vspace{4pt}

  \scriptsize
  \setlength\tabcolsep{3pt}
  \renewcommand{\arraystretch}{1.08}
  \centering
  \begin{tabular}{lcccccccccccccccc}  
    \toprule  
     & \multicolumn{1}{c}{Sp.}
     & \multicolumn{5}{c}{Behavior} &
      \multicolumn{5}{c}{Context} & \multicolumn{5}{c}{Appearance} \\  
    \cmidrule(r){2-2} 
    \cmidrule(r){3-7} \cmidrule(r){8-12} \cmidrule(r){13-17}
     Model   & \multicolumn{1}{R{2cm}}{Species ID}
        & \multicolumn{1}{R{2cm}}{Miscellaneous Behavior}
        & \multicolumn{1}{R{2cm}}{Defensive and \\Survival Behaviors}
        & \multicolumn{1}{R{2cm}}{Cooperative and \\Social Behaviors}
        & \multicolumn{1}{R{2cm}}{Mating, Courtship, \\Reproduction}
        & \multicolumn{1}{R{2cm}}{Feeding and \\Hydration}
        & \multicolumn{1}{R{2cm}}{Human Impact}
        & \multicolumn{1}{R{2cm}}{Miscellaneous \\Context}
        & \multicolumn{1}{R{2cm}}{Animal Structures \\and Habitats}
        & \multicolumn{1}{R{2cm}}{Parasitism and \\Symbiosis}
        & \multicolumn{1}{R{2cm}}{Collected \\Specimens}
        & \multicolumn{1}{R{2cm}}{Tracking and \\Identification}
        & \multicolumn{1}{R{2cm}}{Health and \\Disease}
        & \multicolumn{1}{R{2cm}}{Unique appearances or morphs}
        & \multicolumn{1}{R{2cm}}{Sex identification}
        & \multicolumn{1}{R{2cm}}{Life Cycle and \\Development} \\
    \cmidrule(r){1-1} \cmidrule(rl){2-2} \cmidrule(rl){3-3} \cmidrule{4-4} \cmidrule(rl){5-5}
       \cmidrule(rl){6-6} \cmidrule{7-7} \cmidrule(rl){8-8} \cmidrule(rl){9-9} \cmidrule{10-10}
       \cmidrule(rl){11-11} \cmidrule{12-12} \cmidrule(rl){13-13} \cmidrule{14-14}
      \cmidrule(rl){15-15} \cmidrule(r{1pt}){16-16} \cmidrule(l{1pt}r{1pt}){17-17}
bioclip                   & \textbf{44.8} & 41.2 & 22.9 & 26.4 & 29.4 & 16.7 & 22.2 & 34.6 & 32.5 & 29.4 & 39.9 & 23.7 & 32.2 & 15.9 & \textbf{39.6} & 34.9 \\
rn50                      & 36.6 & 50.8 & 26.0 & 31.7 & 28.4 & 21.6 & 28.4 & 37.1 & 30.6 & 28.9 & 36.4 & 22.4 & 33.1 & 32.7 & 27.5 & 23.7 \\
wildclip-t1t7-lwf         & 40.6 & 47.9 & 33.4 & 31.5 & 32.4 & 19.2 & 38.6 & 35.6 & 23.9 & 25.9 & 31.3 & 29.2 & 31.1 & 30.2 & 32.0 & 27.4 \\
wildclip-t1               & 36.0 & 50.4 & 33.4 & 33.4 & 31.5 & 19.2 & 38.8 & 38.2 & 30.0 & 26.2 & 32.7 & 26.5 & 33.0 & 27.1 & 32.2 & 24.4 \\
vit-b-32                  & 35.3 & 54.8 & 24.1 & 33.2 & 29.5 & 26.0 & 35.0 & 36.0 & 28.9 & 31.9 & 36.1 & 25.8 & 29.4 & 30.1 & 26.8 & 23.6 \\
vit-b-16                  & 36.7 & 52.3 & 29.9 & 35.6 & 30.5 & 23.9 & 39.3 & 35.0 & 27.8 & 27.1 & 35.4 & 36.3 & 31.9 & 37.0 & 28.0 & 22.7 \\
rn50x16                   & 40.0 & 55.7 & 37.8 & 34.5 & 32.7 & 31.6 & 37.6 & 41.9 & 36.5 & 28.8 & 40.2 & 39.9 & 36.5 & 40.5 & 27.9 & 26.0 \\
vit-l-14                  & 38.2 & 47.8 & 36.3 & 39.9 & 36.8 & 29.4 & 45.7 & 40.1 & 34.3 & 27.8 & 30.5 & 45.2 & 35.8 & 46.6 & 27.9 & 30.8 \\
vit-b-16-dfn              & 33.3 & 45.6 & 33.7 & 33.5 & 33.3 & 33.4 & 38.3 & 33.6 & 27.0 & 27.8 & 41.3 & 32.6 & 28.6 & 35.2 & 24.2 & 24.1 \\
vit-l-14-dfn              & 36.1 & 51.1 & 41.7 & 41.5 & 44.8 & 33.2 & 45.5 & 41.0 & 30.3 & 38.6 & 42.6 & 41.2 & 30.9 & 40.4 & 27.8 & 21.1 \\
siglip-vit-l-16-384       & 35.7 & \textbf{70.2} & 46.1 & 56.6 & \textbf{57.9} & 37.0 & 54.4 & \textbf{50.5} & 37.6 & \textbf{40.2} & 52.0 & 65.2 & \textbf{42.0} & 40.6 & 33.2 & \textbf{41.9} \\
vit-h-14-378              & 31.2 & 60.7 & \textbf{57.2} & 51.4 & 52.3 & 39.0 & 49.1 & 39.3 & 33.1 & 38.8 & 43.5 & 60.6 & 34.6 & 40.3 & 27.2 & 30.2 \\
siglip-so400m-14-384      & 44.6 & 68.9 & 45.7 & \textbf{59.1} & 54.6 & \textbf{43.7} & \textbf{58.9} & 50.1 & \textbf{38.4} & 39.9 & \textbf{61.5} & \textbf{77.1} & 39.9 & \textbf{61.4} & 32.1 & 40.6 \\
\midrule
blip2-flan-t5-xxl & 34.2 & 61.8 & 54.8 & 47.9 & 43.1 & 33.1 & 61.3 & 45.5 & 33.8 & 31.5 & 40.5 & 46.9 & 33.8 & 28.4 & 30.0 & 26.1 \\
instructblip-flan-t5-xxl & 37.1 & 59.7 & 52.5 & 49.7 & 45.2 & 37.3 & 71.3 & 52.8 & 32.9 & 30.4 & 46.7 & 62.7 & 40.1 & 36.8 & 31.6 & 25.4 \\
paligemma-3b-mix-448 & 35.8 & 69.1 & 53.2 & 53.7 & 51.9 & 44.3 & 60.6 & 49.2 & 35.5 & 35.0 & \textbf{67.9} & 75.6 & 44.5 & 29.2 & 37.9 & 28.8 \\
llava-1.5-13b & 35.0 & 57.4 & 50.6 & 48.5 & 48.1 & 37.9 & 65.1 & 46.7 & 38.2 & 34.0 & 39.6 & 71.6 & 35.6 & 29.8 & 29.8 & 28.0 \\
llava-v1.6-mistral-7b & 33.4 & 64.9 & 40.0 & 46.3 & 42.6 & 33.2 & 62.8 & 49.3 & 30.1 & 44.0 & 46.3 & 63.2 & 33.7 & 38.2 & 27.9 & 24.9 \\
llava-v1.6-34b & 37.5 & 68.0 & 51.9 & 53.1 & 44.4 & 41.7 & 62.9 & 53.8 & 37.7 & 41.1 & 61.0 & 70.0 & 42.4 & 44.1 & 34.6 & 29.4 \\
VILA1.5-13B & 34.8 & 65.8 & 55.8 & 53.5 & 53.8 & 32.5 & 74.1 & 60.8 & 33.7 & 44.8 & 51.2 & 66.6 & 40.0 & 43.9 & 30.1 & 26.8 \\
VILA1.5-40B & 37.0 & 64.9 & 56.7 & \textbf{61.9} & 52.3 & 46.4 & 71.3 & 63.2 & 44.6 & 51.5 & 61.1 & \textbf{77.6} & 48.5 & \textbf{46.7} & 31.7 & 35.4 \\
\midrule
GPT-4V  & 39.0 & 67.9 & 53.4 & 52.1 & 52.0 & 47.0 & 62.5 & 54.5 & 46.1 & 42.3 & 47.3 & 52.8 & 46.9 & 42.1 & 34.2 & 33.0 \\
GPT-4o      & \textbf{44.3} & \textbf{78.3} & \textbf{70.0} & 61.8 & \textbf{60.5} & \textbf{56.6} & \textbf{82.4} & \textbf{69.2} & \textbf{49.6} & \textbf{64.2} & 67.5 & 72.7 & \textbf{55.6} & 45.6 & \textbf{44.3} & \textbf{41.2} \\
  \bottomrule
  \end{tabular}
  \vspace{-2em}
\end{table}

\subsection{Validation Set Results}
In Table~\ref{tab:supp_fullrank_val_and_test} we compare the results on the validation and test set for \textsc{INQUIRE-Fullrank}. We observe that the scores are broadly similar on both data splits, confirming that the validation split is a representative sample of \inquire{} queries.

\begin{table}[]
    \centering
    \small
    \caption{Comparison of \textsc{INQUIRE-Fullrank} results between the validation set and test set.}
    \label{tab:supp_fullrank_val_and_test}
    \vspace{4pt}
    \renewcommand{\arraystretch}{1.1}
    \begin{tabular}{llcccccc}
        \toprule
         & &  \multicolumn{2}{c}{AP@50} & \multicolumn{2}{c}{nDCG@50} & \multicolumn{2}{c}{MRR} \\  
    \cmidrule(r){3-4} 
    \cmidrule(r){5-6} \cmidrule(r){7-8}
    Dataset & Model & Val & Test & Val & Test & Val & Test \\ \midrule
WildCLIP   & CLIP ViT-B-16          & 6.9 & 7.4 & 14.3 & 16.1 & 0.24 & 0.33 \\
BioCLIP    & CLIP ViT-B-16          & 1.7 & 5.0 & 5.2 & 8.6 & 0.11 & 0.17 \\
OpenAI     & CLIP RN50              & 8.7 & 6.8 & 18.3 & 15.1 & 0.35 & 0.29 \\
OpenAI     & CLIP RN50x16           & 16.8 & 13.6 & 28.0 & 25.5 & 0.47 & 0.48 \\
OpenAI     & CLIP ViT-B-32          & 9.8 & 7.5 & 18.9 & 16.8 & 0.32 & 0.30 \\
OpenAI     & CLIP ViT-B-16          & 11.7 & 10.4 & 23.1 & 20.9 & 0.45 & 0.40 \\
OpenAI     & CLIP ViT-L-14          & 18.4 & 14.4 & 31.5 & 27.1 & 0.47 & 0.46 \\
DFN        & CLIP ViT-B-16          & 16.4 & 15.1 & 29.1 & 28.1 & 0.48 & 0.48 \\
DFN        & CLIP ViT-L-14          & 22.4 & 23.1 & 36.5 & 37.3 & 0.55 & 0.54 \\
DFN        & CLIP ViT-H-14@378      & 35.1 & 33.3 & 51.2 & 48.8 & 0.71 & 0.69 \\
WebLI      & SigLIP ViT-L-16@384    & 32.1 & 31.1 & 47.3 & 46.6 & 0.67 & 0.68 \\
WebLI      & SigLIP SO400m-14@384   & 30.0 & 34.2 & 46.0 & 49.1 & 0.64 & 0.69 \\
\bottomrule
    \end{tabular}
\end{table}

\section{Computational Efficiency} \label{appendix-computational-efficiency}

\begin{figure}[h]
    \centering
    \includegraphics[height=4.1cm]
    {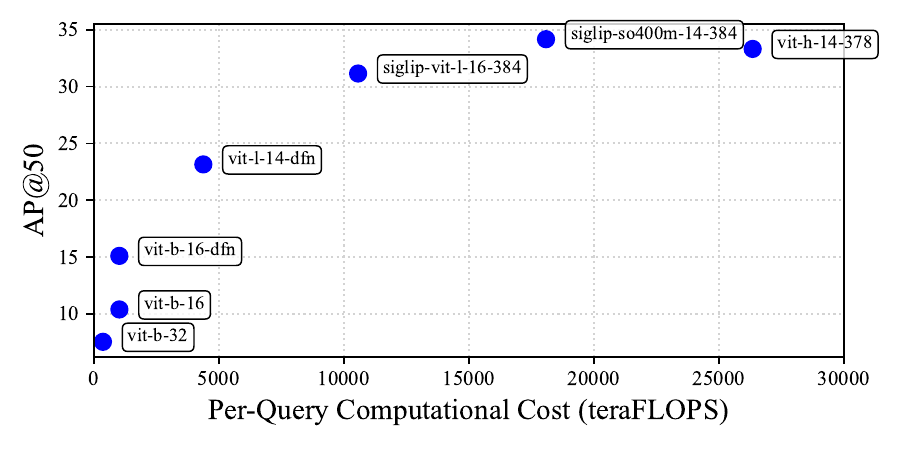} 
    \includegraphics[height=4.1cm]{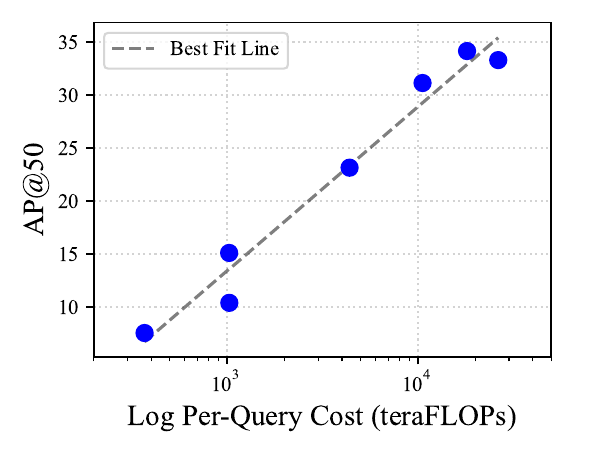}
    \caption{Computation cost of different CLIP models plotted against their performance on \inquire{} full dataset retrieval.}
    \label{fig:computational_efficiency}
\end{figure}

\textbf{Embedding Generation.} 
The computational efficiency of a retrieval method is key to its real-world viability. In Figure~\ref{fig:computational_efficiency} we estimate the computational cost for selected CLIP retrieval methods. 
Here, the computational cost represents  the total computational cost of generating all CLIP embeddings (the per-inference cost is provided by OpenCLIP~\cite{ilharco_gabriel_2021_5143773}), and dividing by 250, the number of queries. However, we note that in practice, once all image embeddings are pre-computed and stored in an efficient nearest-neighbors index (\eg Faiss~\cite{douze2024faiss}), each query takes milliseconds and thus its search cost is near-zero. The only significant computational cost will be that of performing inference on the  query via the text encoder.

\textbf{Scaling Laws.} Figure~\ref{fig:computational_efficiency}-left also shows diminishing returns in AP@50 as the model size, and thus the computational cost, increases. When we plot the same data using a log-scaled x-axis in Figure~\ref{fig:computational_efficiency}-right, we observe a roughly linear trend between the log-scaled computational cost and the AP@50. While further study is required to fully characterize this particular trend, this result shows evidence of power law scaling similar to other machine learning tasks~\cite{cherti2023reproducible}.

\textbf{Computational Resources Used.}
All experiments were performed on A100 GPUs.

\section{Evaluation Metrics}\label{appendix-metrics}

\paragraph{Average Precision at k.} 
Average Precision (AP) is a well-known metric computed by taking the weighted mean of precision scores at a set of thresholds. This metric has been adapted to the retrieval setting, where it possible to calculate the Average Precision at k (AP@k) among just the top k retrieved items. Since calculating AP@k requires both the relevance and position of the top k items, AP@k may be prefered over Precision at k (P@k) which does not use position. A number of AP@k variants have been proposed~\cite{baeza1999modern, hawking2001measuring, voorhees2005trec}, taking the general form:

\begin{equation}
    AP@k = \frac{\sum_{i=1}^k P@i \cdot rel(i)}{NF}
\end{equation}

where $Pr@i$ is the precision at $i$ (\ie among the first $i$ items), $rel(i) \in \{0,1\}$ is the binary relevance score, and $NF$ is a normalization factor. 

In a typical implemenation of AP we would see $NF = r$, the total number of relevant items in the top $k$. However in a retrieval setting with a total of $R$ relevant items, this normalization technique creates a problematic and unintuitive situation where promoting an item into the top k retrievals can decrease the score. 

In particular, consider the situation where we have 100 images of which 2 are relevant and 98 are not relevant. Using a normalization factor of $NF=R$, we measure AP@5 for the following two top-5 retrievals:

\begin{enumerate}
    \item Ordered retrieval relevance: (1, 0, 0, 0, 0)  $\implies$ AP@5 = 1
    \item Ordered retrieval relevance: (1, 0, 0, 0, 1)  $\implies$ AP@5 = 0.7 
\end{enumerate}

We observe that promoting a relevant item into the top 5 resulted in a decreased AP@5, which is undesirable. Our criteria for an AP@ metric is that (1) the measure strictly increases whenever a relevant document is promoted into the top-k, and (2) the has a full range of 0 to 1. Of the range of proposed AP@k variants~\cite{baeza1999modern, hawking2001measuring, voorhees2005trec}, just \cite{voorhees2005trec} meets our desired criteria This modified average precision normalizes using $min(k,R)$. In the case above, we now have $NF = min(k, R) = min(5, 2) = 2$, yielding:

\begin{enumerate}
    \item Ordered retrieval relevance: (1, 0, 0, 0, 0)  $\implies$ AP@5 = 0.5
    \item Ordered retrieval relevance: (1, 0, 0, 0, 1)  $\implies$ AP@5 = 0.7
\end{enumerate}

Our end-to-end retrieval evaluations use AP@k with this desirable normalization factor of $NF = min(k, R)$. Since the reranking challenge evaluates solely using the fixed set, the normalization factor for this challenge is always $r$, the number of relevant items within the top k. 

For further discussion of Average Precision at k, we refer readers to \cite{Craswell2009}.

\textbf{nDCG}. Normalized discounted cumulative gain is a weighted ranking metric that considers the relative ordering of the retrieved items. To compute nDCG@k for a single query, first we compute the discounted cumulative gain at k (DCG@K):

\begin{equation}
    DCG_k = \sum_{i=1}^k\frac{rel(i)}{\log_2(i+1)}
\end{equation}
where $rel(i) \in \{0, 1\}$ is the binary relevance score for the $i$th retrieved item. Then, we define the ideal DCG at k (IDCG@k) as the maximum achievable DCG@k:
\begin{equation}
    IDCG_k = \sum_{i=1}^{\min(k, R)}\frac{rel(i)}{\log_2(i+1)}
\end{equation}
where $R$ is the total number of relevant items for the query. Finally, we can compute $nDCG@k$ as
\begin{equation}
    nDCG_k = \frac{DCG_k}{IDCG_k}
\end{equation}
where the normalization by $IDCG@k$ allows $nDCG_k$ to range fully between the interval 0 to 1.

\textbf{MRR}. Mean reciprocal rank is a measure for the rank of the first correct retrieval. MRR can be computed as
\begin{equation}
    MRR = \frac{1}{Q}\sum_{i=1}^Q\frac{1}{rank(i)}
\end{equation}
where $Q$ is the number of queries, and $rank(i)$ gives the rank of the first relevant retrieval for the $i$th query (1 for 1st position, 2 for 2nd position, \etc). If no relevant retrievals are present in the retrieved list, we let $rank(i) = \infty$, \ie $1/rank(i) = 0$.

\section{iNat24 Image Collection and INQUIRE Annotation Protocol}\label{appendix-data-annotation}
In this sections we describe in detail our data collection protocol for collecting the iNat24 dataset and annotating the \inquire{} benchmark.

\subsection{iNat24 Dataset Curation}\label{appendix-data-annotation-curation}
We follow a similar paradigm used to organize the iNaturalist Competition Datasets from 2017~\cite{van2018inaturalist}, 2018~\cite{iNatCompGithub}, 2019~\cite{iNatCompGithub}, and 2021~\cite{van2021benchmarking}. 
For the 2024 version we start from an iNaturalist observation database export generated on 2023-12-30. 
Observations are then filtered to include only those observed in the years 2021, 2022, or 2023. 
This ensures the images in iNat24 are unique and do not overlap with images from prior dataset versions (\eg iNat21~\cite{van2021benchmarking} only contains images up until September 2020). 
To utilize the iNat21 taxonomy (for easy compatibility with that dataset) we detect taxonomic changes between the iNat21 taxonomy and the iNaturalist taxonomy included in the 2023-12-30 database export. 
We then modify species labels (where necessary) so that observations conform to the iNat21 taxonomy. 
Some of these taxonomic changes can be quite complicated (splits, merges, \etc) resulting in cases where an iNat21 species is no longer valid, however we are able to recover 9,959 out of the original 10,000 species from iNat21.  
We then filter to include observations exclusively from the iNat21 taxonomy.  
Additional filtering ensures that all observations have valid metadata (\ie location and time information) and that associated image files are not corrupted.
These steps result in a candidate set of 33M observations to sample from to build the iNat24 dataset. 

Our process of selecting the set of images to include for each species in the iNat24 dataset deviates from the prior dataset building schemes~\cite{van2018inaturalist, van2021benchmarking}. 
Random sampling of observations, or even random sampling from unique users, generates collections of images that are biased towards North America and Europe. 
To decrease this bias we sample from spatio-temporal clusters of ``observations groups''.  
Observation groups are formed by grouping observations together if they are observed on the same day within 10km of each other, regardless of the observer. 
When sampling observations for a species, we cluster their associated observation groups using a spatio-temporal distance metric and then sample one observation per cluster in a round-robin fashion until we hit a desired sample size. 
When sampling within a cluster, we prioritize novel observation groups and novel users.    
We sample at most 550 observations per species to include in iNat24. 
This sampling process results in a total of 4,816,146 images for 9,959 species. 

Unlike previous versions of the iNaturalist dataset, we performed one final round of filtering to remove images that are inappropriate for a research dataset or not relevant for the query. 
We use the INQUIRE annotation process to find images containing human faces, personally identifiable information, ``empty'' images, images of spectrograms, \etc. We additionally run a RetinaFace~\cite{deng2020retinaface} Resnet50 face detection model across the entire dataset, and manually inspect all high confidence predictions.
In total this filtered out an additional 2,603 images. 
The final dataset contains 4,813,543 images for 9,959 species.

The iNat24 dataset does not have a validation or test split, \ie all observations are assigned to the train split. 
The validation and test splits can be used from the iNat21 dataset to benchmark classification performance. 
As in previous years, we keep only the primary image for each observation, and resize all images to have a max dimension of 500px on the longest side. 
All images have three channels and are stored as jpegs. 
We provide location, time, attribution, and licensing information in the associated json file.

\subsection{Data Annotation}
\label{appendix-data-annotation-general}

\begin{figure}[t]
    \centering
    \includegraphics[width=\textwidth]{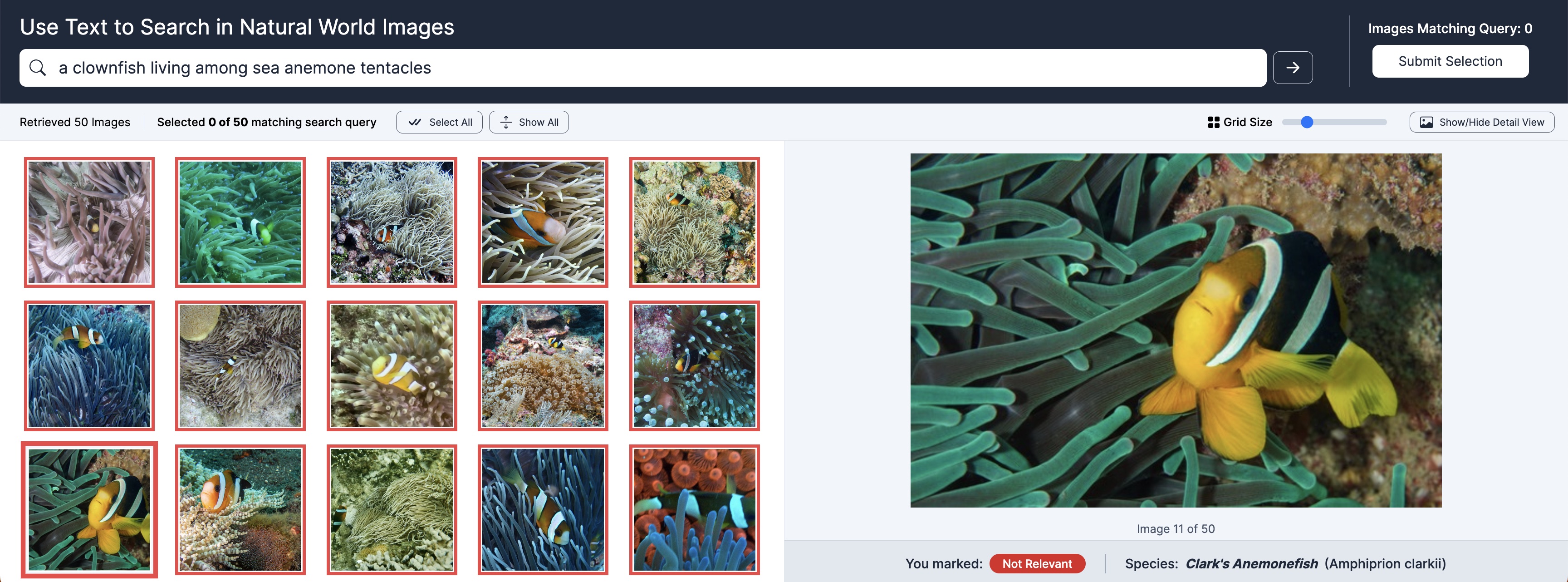}
    \caption{Here we display a screen shot of the online annotation tool we developed for annotation. The tool supports CLIP similarity search and species filtering. 
    }
    \label{fig:tool}
\end{figure}

Image annotation was performed by a carefully selected team of paid MSc students or equivalent, many with expertise in ecology allowing for labeling of difficult queries. 
Annotators were instructed to label all candidate images as either relevant (\ie positive match) or not relevant (\ie negative match) to the query, and to mark an image as not relevant if there was reasonable doubt as to its relevance. 
At this stage, queries that were deemed very easy, not comprehensively labeled, or otherwise not possible to label were excluded from the benchmark.

The annotation itself is performed using a custom interface that we developed that shows the top retrievals given a text query and optionally allows the user to filter based on the species label. 
A screen shot of the tool is displayed in Figure~\ref{fig:tool}. 
The retrievals are ordered by CLIP ViT-H/14~\cite{fang2023data} similarity to the query text. 
Annotators generally label at least 500 images per query.

We comprehensively labeled the dataset primarily through the use of species filters for a single or a group of species. For example, to thoroughly label the query \textit{``Black Skimmer performing skimming"}, a single species filter (Black Skimmer) was utilized while for the query \textit{``flamingo standing on one leg"}, four different species filters were needed to account for all the flamingo species included in iNat24 (Lesser Flamingo, Chilean Flamingo, Greater Flamingo, and American Flamingo). Using species filters in this way allows us to sufficiently reduce the search space for these queries to comprehensively label iNat24 for all possible matches.

When a query corresponds to a very large number of species, or no species in particular (\eg \textit{``an image containing a photographic reference scale with a color swatch''}), we label using just the top CLIP retrievals without any filters. In this case, we tend to label a significantly larger number of images, and we label until at least 100 images in a row are negative indicating that the set of positives has been exhausted. If this condition is not met after at a large number of labels, or the annotator otherwise believes that comprehensive labeling is not possible, we do not use the query. We note that the quality of our comprehensive labeling in this case is limited by the CLIP model's ability to surface relevant positives, so any missed positives with lower relevance score could be left unlabeled. This affects only 12 of our 250 queries for which we label without species filters.
This is a primary motivator behind the large number of images labeled per query (\ie >500). However, if there were indeed missed positives, then we would expect the CLIP ViT-H/14 model used for labeling to perform unexpectedly well, as higher quality models that surface missed positive image would be penalized as these would be considered negative at evaluation time. Yet, our evaluation in Table~\ref{tab:zero_shot_results} shows that SigLIP, which on OpenCLIP's~\cite{ilharco_gabriel_2021_5143773} retrieval evaluation performs only marginally better than CLIP ViT-H/14, achieves a comparable score. This result suggests that our dataset does not suffer from a significant missed positive issue. 

Creating \inquire{} involved labeling 194,334 images, of which 32,696 were relevant to their queries. Labeling took place over a total of about 180 hours, so the average time spent labeling is 43 minutes per query or 3.3 seconds per image.

\subsection{Data Format and Structure}
\textbf{iNat24}. iNat24 is provided as a metadata file and a tar file containing all images. The metadata file is given in the commonly used JSON COCO format. The information in this metadata file includes each image's ID, file path, width, height, image license, rights holder, taxonomic classification, latitude, longitude, location uncertainty, and date.

\textbf{INQUIRE}. The \inquire{} benchmark is provided as a two CSV files. The first is a list of queries, where each row includes fields for the query id, query text, organism category, query category type, and query category. The second file is a list of annotations, where row corresponds includes fields for the query id, image id, and relevance label. The image id can be matched to the iNat24 metadata to get additional information mentioned above, such as the taxonomy, date, and geographic location.

\subsection{Ethical Considerations}

\textbf{Copyright and Licensing}. We adhere strictly to copyright and licensing regulations. All images included in the dataset fall under a license allowing copying and redistribution. In particular, all images are licensed under one of the following: 
CC BY 4.0, CC BY-NC 4.0, CC BY-NC-ND 4.0, CC BY-NC-SA 4.0, CC0 1.0, CC BY-ND 4.0, or CC BY-SA 4.0.

\textbf{Data Privacy and Safety}. Although users approved all images considered for research use, we take further steps to ensure data privacy and safety. We filter all images for content that is contains personally identifiable information or images of people. 
We do not exclude most images containing gore, as these are often ecologically relevant, \eg using image of road-killed animals to asses impacts of roads on biodiversity.

\textbf{Violations of Rights}. We respect the rights of iNaturalist community volunteer observers by constructing iNat2024 using only images and metadata appropriately licensed by their respective creators for copying, distribution, and non-commercial research use. Nevertheless, we bear responsibility in case of a violation of rights.

\textbf{Participant Risks}.  We received internal ethical approval for our query collection and data labeling (Edinburgh Informatics Ethics Review Panel 951781 and MIT Committee on the Use of Humans as Experimental Subjects Protocol 2404001276). 

\subsection{Participant Compensation} \label{appendix-data-annotation-spending}
We hired annotators at the equivalent of \$15.50 per hour and spent a total of \$2325 on annotation. %

\subsection{Annotation Instructions}\label{appendix-labeler-instructions}
The instructions provided to annotators are included below.

\includepdf[pages={1-15}, nup=2x4, scale=0.8, frame=true]{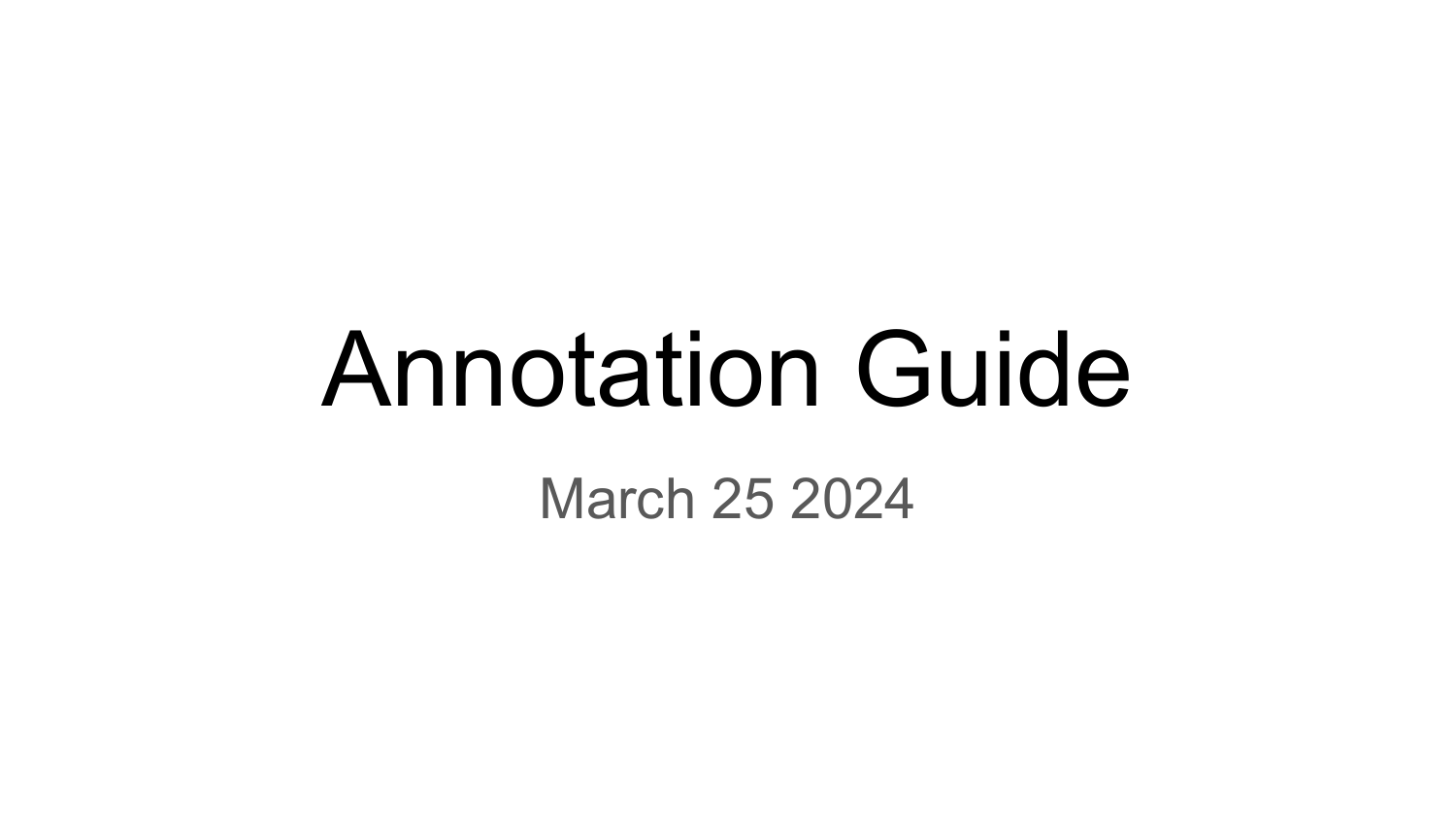}

\section{Multi-Modal Model Prompting Details}\label{appendix-lmm-prompting}
We include the various prompts used in our evaluation of large multimodal models in Table~\ref{tab:multimodal_queries}. 
We note that while we aim to keep the prompt broadly the same across models, they are ultimately different due to different prompting requirements for each model. The proprietary models (GPT-4V and GPT-4o) were queries on October 14, 2024.

\begin{table}[ht]
    \centering
    \caption{Format of the text prompts used by the large multimodal models.}
    \vspace{5pt}
\renewcommand{\arraystretch}{1.1}
\begin{tabular}{lp{10cm}}
    \toprule
    BLIP-2 &     \texttt{Does this picture show \{query\}?$\backslash$nAnswer the question with either "Yes" or "No" and nothing else.}   \\ \midrule
    InstructBLIP & \texttt{Does this picture show \{query\}?$\backslash$nAnswer the question with either "Yes" or "No" and nothing else.} \\ \midrule
    LLaVA-v1.5 & \texttt{USER: <image>$\backslash$nDoes this picture show \{query\}? Answer the question with either "Yes" or "No" and nothing else.$\backslash$nASSISTANT:} \\ \midrule
    LLaVA-v1.6-7b & \texttt{[INST] <image>$\backslash$nDoes this picture show \{query\}? Answer the question with either "Yes" or "No" and nothing else. [/INST]} \\ \hline
    LLaVA-v1.6-34b & \texttt{<|im\_start|>system$\backslash$nAnswer the questions.<|im\_end|>
    <|im\_start|>user
    $\backslash$n<image>$\backslash$nDoes this image show "\{query\}"? Answer the question with either "Yes" or "No".
    <|im\_end|><|im\_start|>answer$\backslash$n} \\ \hline
    PaliGemma & \texttt{Q: Does this picture show \{query\}? Respond with yes or no.$\backslash$nA:} \\ \hline
    VILA-v1.5-13B & \texttt{<image>$\backslash$n Does this picture show \{query\}?$\backslash$nAnswer the question with either "Yes" or "No" and nothing else.} \\ \hline
    VILA-v1.5-40B & \texttt{<image>$\backslash$n Does this picture show \{query\}?$\backslash$nAnswer the question with either "Yes" or "No" and nothing else.} \\ \hline
    GPT-4V & \texttt{Does this picture show exactly "\{query\}"?$\backslash$nAnswer the question with either "Yes" or "No" and nothing else.} \\ \hline
    GPT-4o & \texttt{Does this picture show exactly "\{query\}"?$\backslash$nAnswer the question with either "Yes" or "No" and nothing else.} \\
    \bottomrule
\end{tabular}
\label{tab:multimodal_queries}
\end{table}

\section{Full List of \inquire{} Queries}
\label{appendix-additional-examples}
Table~\ref{tab:sup_full_query_list} lists all \inquire{} queries.

{
\renewcommand{\arraystretch}{1.18}
\small
\begin{longtable}{| p{.45\textwidth}  l  l |} 
\caption{\textsc{Inquire} includes 250 queries across a range of categories. This table lists all 250 queries along with the supercategory and category that they belong to.}\label{tab:sup_full_query_list}
\endfirsthead
\endhead
\hline
\textbf{Query} & \textbf{Supercategory} & \textbf{Category} \\ \hline
A By-the-Wind Sailor washed up on a beach & Appearance & Health and Disease \\ \hline 
A beached Portuguese Man o' War & Appearance & Health and Disease \\ \hline 
Dead hog-nosed skunk & Appearance & Health and Disease \\ \hline 
Moorish Gecko with regenerated tail & Appearance & Health and Disease \\ \hline 
Redwood tree with fire scars & Appearance & Health and Disease \\ \hline 
Wolf spider with limb loss & Appearance & Health and Disease \\ \hline 
beached orca & Appearance & Health and Disease \\ \hline 
black knot caused by a fungal pathogen & Appearance & Health and Disease \\ \hline 
common murre beached carcass & Appearance & Health and Disease \\ \hline 
elk with hair loss & Appearance & Health and Disease \\ \hline 
fire pink with dark-colored anthers & Appearance & Health and Disease \\ \hline 
moose with hair loss & Appearance & Health and Disease \\ \hline 
sick cassava plant & Appearance & Health and Disease \\ \hline 
southern aligator lizard with cut tail & Appearance & Health and Disease \\ \hline 
turtle roadkill & Appearance & Health and Disease \\ \hline 
A cicada in the process of shedding its exoskeleton & Appearance & Life Cycle and Development \\ \hline 
Blackening Waxcap mushroom turned fully black & Appearance & Life Cycle and Development \\ \hline 
Caribou without palmate antlers & Appearance & Life Cycle and Development \\ \hline 
Cart-Rut Shell snail egg mass & Appearance & Life Cycle and Development \\ \hline 
Cooper's Hawk in adult plumage & Appearance & Life Cycle and Development \\ \hline 
Lilac Bonnet with edges turnt up revealing its gills & Appearance & Life Cycle and Development \\ \hline 
Luna Moth missing long twisted tails on the hindwings & Appearance & Life Cycle and Development \\ \hline 
Male Blue-black Grassquit in patchy blue-black and brown plumage & Appearance & Life Cycle and Development \\ \hline 
Swallowtail butterfly caterpillar camouflaged as bird droppings & Appearance & Life Cycle and Development \\ \hline 
a bag-shelter moth's silken bag nest on a tree & Appearance & Life Cycle and Development \\ \hline 
a bag-shelter moth's silken bag nest on the ground near a plant & Appearance & Life Cycle and Development \\ \hline 
breeding adult Black-bellied Plover & Appearance & Life Cycle and Development \\ \hline 
bullhead shark egg case & Appearance & Life Cycle and Development \\ \hline 
caterpillar of Common Buckeye butterfly & Appearance & Life Cycle and Development \\ \hline 
caterpillar of Gulf Fritillary butterfly & Appearance & Life Cycle and Development \\ \hline 
caterpillar of polyphemus moth & Appearance & Life Cycle and Development \\ \hline 
laysan albatross mostly in dark mottled brown plumage & Appearance & Life Cycle and Development \\ \hline 
monkey slug caterpillar & Appearance & Life Cycle and Development \\ \hline 
A female pheasant & Appearance & Sex identification \\ \hline 
European Stag Beetle lacking mandibles & Appearance & Sex identification \\ \hline 
Male Xanthagrion erythroneurum & Appearance & Sex identification \\ \hline 
Male crimsonband wrasse & Appearance & Sex identification \\ \hline 
Rusty tussock moth adult female & Appearance & Sex identification \\ \hline 
Splendid fairywren without blue body coloration & Appearance & Sex identification \\ \hline 
adult male Misumena vatia & Appearance & Sex identification \\ \hline 
female beautiful demoiselle & Appearance & Sex identification \\ \hline 
female or immature evening grosbeak & Appearance & Sex identification \\ \hline 
fiddler crab with an oversized chela & Appearance & Sex identification \\ \hline 
male Northern Elephant Seal & Appearance & Sex identification \\ \hline 
male ruby-throated hummingbird in flight & Appearance & Sex identification \\ \hline 
male velvet ant & Appearance & Sex identification \\ \hline 
Black Skimmer with a leg band & Appearance & Tracking and Identification \\ \hline 
Cheetah with radio collar & Appearance & Tracking and Identification \\ \hline 
North Island robin or South Island robin with leg bands & Appearance & Tracking and Identification \\ \hline 
Rhino with ear notches & Appearance & Tracking and Identification \\ \hline 
Tagged swan & Appearance & Tracking and Identification \\ \hline 
a bighorn sheep with a tracking collar around its neck & Appearance & Tracking and Identification \\ \hline 
a lion with a collar around its neck & Appearance & Tracking and Identification \\ \hline 
an image showing a humpback whale fluke with clearly identifiable markings & Appearance & Tracking and Identification \\ \hline 
Axanthism in a green frog (Lithobates clamitans) & Appearance & Unique appearances or morphs \\ \hline 
Fly Agaric in yellow form & Appearance & Unique appearances or morphs \\ \hline 
House finch with yellow pigmentation & Appearance & Unique appearances or morphs \\ \hline 
Melanistic jaguar & Appearance & Unique appearances or morphs \\ \hline 
Melanistic leopard & Appearance & Unique appearances or morphs \\ \hline 
Strawberry poison-dart frog with the "blue jeans" color morph & Appearance & Unique appearances or morphs \\ \hline 
Strawberry poison-dart frog with the "la gruta" color morph from Isla Colon & Appearance & Unique appearances or morphs \\ \hline 
Tropical Orb Weaver with multiple white spots in the abdominal dorsum & Appearance & Unique appearances or morphs \\ \hline 
albino american robin & Appearance & Unique appearances or morphs \\ \hline 
dark-colored Dusky Leaf-monkey with orange-colored young & Appearance & Unique appearances or morphs \\ \hline 
eastern gray squirrel displaying melanistic pelage & Appearance & Unique appearances or morphs \\ \hline 
fire salamander with a barred color pattern & Appearance & Unique appearances or morphs \\ \hline 
A close-up of an ant carrying a leaf & Behavior & Cooperative and Social Behaviors \\ \hline 
A meadowlark vocalizing & Behavior & Cooperative and Social Behaviors \\ \hline 
Emergence of large colony of mexican free-tailed bats & Behavior & Cooperative and Social Behaviors \\ \hline 
Fur seal mother with pup & Behavior & Cooperative and Social Behaviors \\ \hline 
Sage Thrasher vocalizing & Behavior & Cooperative and Social Behaviors \\ \hline 
Scorpion with young on its back & Behavior & Cooperative and Social Behaviors \\ \hline 
Two giraffes & Behavior & Cooperative and Social Behaviors \\ \hline 
a herd of more than 10 impalas & Behavior & Cooperative and Social Behaviors \\ \hline 
a sandhill crane couple with their chicks & Behavior & Cooperative and Social Behaviors \\ \hline 
canada geese flying in v-formation & Behavior & Cooperative and Social Behaviors \\ \hline 
cheetah with cubs & Behavior & Cooperative and Social Behaviors \\ \hline 
couple of black-bellied whistling ducks with their youngs sharing parenting duties & Behavior & Cooperative and Social Behaviors \\ \hline 
grebe with babies on its back & Behavior & Cooperative and Social Behaviors \\ \hline 
jungle babblers allopreening & Behavior & Cooperative and Social Behaviors \\ \hline 
macaques engaging in mutual grooming behavior & Behavior & Cooperative and Social Behaviors \\ \hline 
male Red-winged Blackbird vocalizing & Behavior & Cooperative and Social Behaviors \\ \hline 
mountain goat climbing rocky outcrops with its young & Behavior & Cooperative and Social Behaviors \\ \hline 
photo with two or more adult male lions & Behavior & Cooperative and Social Behaviors \\ \hline 
picture showing more than fifty Mediterranean Acrobat Ants & Behavior & Cooperative and Social Behaviors \\ \hline 
picture showing more than fifty Velvety Tree Ants & Behavior & Cooperative and Social Behaviors \\ \hline 
A mongoose standing upright alert & Behavior & Defensive and Survival Behaviors \\ \hline 
Gazelle being vigilant/looking around & Behavior & Defensive and Survival Behaviors \\ \hline 
Inflated pufferfish & Behavior & Defensive and Survival Behaviors \\ \hline 
Red-capped Plover performing broken wing distraction & Behavior & Defensive and Survival Behaviors \\ \hline 
antelopes head-butting & Behavior & Defensive and Survival Behaviors \\ \hline 
everted osmeterium & Behavior & Defensive and Survival Behaviors \\ \hline 
hognose snake playing dead & Behavior & Defensive and Survival Behaviors \\ \hline 
killdeer feigning injury & Behavior & Defensive and Survival Behaviors \\ \hline 
moray eel with open mouth poking head out of burrows or crevices & Behavior & Defensive and Survival Behaviors \\ \hline 
vigilant prairie dog stands guard & Behavior & Defensive and Survival Behaviors \\ \hline 
white-tailed deer flagging its tail & Behavior & Defensive and Survival Behaviors \\ \hline 
A male and female cardinal sharing food & Behavior & Feeding and Hydration \\ \hline 
Butterflyfish feeding on brain coral & Behavior & Feeding and Hydration \\ \hline 
Elephants at a watering hole & Behavior & Feeding and Hydration \\ \hline 
Leafhopper Assassin Bug predating lady beetle & Behavior & Feeding and Hydration \\ \hline 
Milkweed Assassin Bug predating a bee or wasp & Behavior & Feeding and Hydration \\ \hline 
Northern Mockingbird carrying out its food & Behavior & Feeding and Hydration \\ \hline 
Surgeonfish grazing on algae & Behavior & Feeding and Hydration \\ \hline 
Yellow-faced Honeyeater in birdbath & Behavior & Feeding and Hydration \\ \hline 
black-winged kite carrying prey in its talons & Behavior & Feeding and Hydration \\ \hline 
blue jay eating whole peanuts & Behavior & Feeding and Hydration \\ \hline 
feral cat with prey & Behavior & Feeding and Hydration \\ \hline 
fruit bat eating fruit upside down & Behavior & Feeding and Hydration \\ \hline 
great golden digger wasp carrying an orthopteron & Behavior & Feeding and Hydration \\ \hline 
hyenas eating a kill & Behavior & Feeding and Hydration \\ \hline 
parrotfish feeding & Behavior & Feeding and Hydration \\ \hline 
puffins carrying food & Behavior & Feeding and Hydration \\ \hline 
red-tailed hawk perched on a utility pole & Behavior & Feeding and Hydration \\ \hline 
water snake feeding on fish & Behavior & Feeding and Hydration \\ \hline 
A peacock male displaying its feathers & Behavior & Mating, Courtship, Reproduction \\ \hline 
Alligator lizards mating & Behavior & Mating, Courtship, Reproduction \\ \hline 
Elk bugling during the rut & Behavior & Mating, Courtship, Reproduction \\ \hline 
Haliaeetus eagles in aerial cartwheeling flight with locked talons & Behavior & Mating, Courtship, Reproduction \\ \hline 
Hübner's Wasp Moth mating & Behavior & Mating, Courtship, Reproduction \\ \hline 
Male chorus frog with inflated vocal sacs & Behavior & Mating, Courtship, Reproduction \\ \hline 
Nursery Web Spider carrying egg sac & Behavior & Mating, Courtship, Reproduction \\ \hline 
Water Frogs in amplexus position & Behavior & Mating, Courtship, Reproduction \\ \hline 
male smooth newt with developed crest & Behavior & Mating, Courtship, Reproduction \\ \hline 
pair of great crested grebes potentially performing the weed dance & Behavior & Mating, Courtship, Reproduction \\ \hline 
vervet monkey with visible bright blue scrotal areas & Behavior & Mating, Courtship, Reproduction \\ \hline 
Big Brown Bat roosting & Behavior & Miscellaneous Behavior \\ \hline 
Dolphins performing acrobatics & Behavior & Miscellaneous Behavior \\ \hline 
Eastern Red Bat in flight & Behavior & Miscellaneous Behavior \\ \hline 
Young sea turtles heading towards the ocean & Behavior & Miscellaneous Behavior \\ \hline 
a mud dauber wasp in the process of forming its nest out of mud & Behavior & Miscellaneous Behavior \\ \hline 
elephant covered in mud or dirt & Behavior & Miscellaneous Behavior \\ \hline 
flamingo standing on one leg & Behavior & Miscellaneous Behavior \\ \hline 
A Eurasian Red Squirrel gathering material for its nest & Context & Animal Structures and Habitats \\ \hline 
A case made by a bagworm larva & Context & Animal Structures and Habitats \\ \hline 
A satin bowerbird's bower ornamented with blue objects & Context & Animal Structures and Habitats \\ \hline 
Hamerkop collecting nesting material & Context & Animal Structures and Habitats \\ \hline 
a beaver dam across a stream & Context & Animal Structures and Habitats \\ \hline 
a gall produced by an Acorn Plum Gall Wasp & Context & Animal Structures and Habitats \\ \hline 
an oak gall produced by a spongy oak apple gall wasp & Context & Animal Structures and Habitats \\ \hline 
potter wasp nest & Context & Animal Structures and Habitats \\ \hline 
camera trap photo of a bobcat captured in the nightime & Context & Collected Specimens \\ \hline 
camera trap photo of a springbok drinking water & Context & Collected Specimens \\ \hline 
camera trap photo of a stag red deer & Context & Collected Specimens \\ \hline 
camera trap photo of chital with its head down & Context & Collected Specimens \\ \hline 
measuring the body dimensions of a bee & Context & Collected Specimens \\ \hline 
A coyote resting on human-build structures & Context & Human Impact \\ \hline 
Fishing net on a reef & Context & Human Impact \\ \hline 
Maple tree with signs of tapping & Context & Human Impact \\ \hline 
a hermit crab using plastic waste as a shell & Context & Human Impact \\ \hline 
a possum on a power line & Context & Human Impact \\ \hline 
bird caught in a net & Context & Human Impact \\ \hline 
bird dead in front of a window & Context & Human Impact \\ \hline 
brown bear near vehicle & Context & Human Impact \\ \hline 
elephant near a fence & Context & Human Impact \\ \hline 
human handling a bat with bare hands & Context & Human Impact \\ \hline 
sloth hugging a human-made structure & Context & Human Impact \\ \hline 
A close-up of a Star-nosed Mole's nose showing all appendages of its Eimer's organs & Context & Miscellaneous Context \\ \hline 
A koala that is not in a tree & Context & Miscellaneous Context \\ \hline 
Adult female black widow spider not in a web & Context & Miscellaneous Context \\ \hline 
Exactly five penguins & Context & Miscellaneous Context \\ \hline 
Exactly three zebras standing together in a field & Context & Miscellaneous Context \\ \hline 
Mushrooms growing in a fairy ring formation & Context & Miscellaneous Context \\ \hline 
Octopus out of the water & Context & Miscellaneous Context \\ \hline 
Scarlet Waxy Cap with visible gills & Context & Miscellaneous Context \\ \hline 
a microscopy slide showing the cellular structure of a plant & Context & Miscellaneous Context \\ \hline 
an image containing a photographic reference scale with a color swatch & Context & Miscellaneous Context \\ \hline 
dorsal side of mourning cloak butterfly & Context & Miscellaneous Context \\ \hline 
hot lips plant with blue fruits & Context & Miscellaneous Context \\ \hline 
ventral side of mourning cloak butterfly & Context & Miscellaneous Context \\ \hline 
Channeled Applesnail covered by green algae & Context & Parasitism and Symbiosis \\ \hline 
Chinese Mystery Snail covered by green algae & Context & Parasitism and Symbiosis \\ \hline 
Mexican grass-carrying wasp visiting a purple flower & Context & Parasitism and Symbiosis \\ \hline 
Sea turtle with algae on its shell & Context & Parasitism and Symbiosis \\ \hline 
Sharks with remoras attached & Context & Parasitism and Symbiosis \\ \hline 
Zebra and wildebeest grazing together & Context & Parasitism and Symbiosis \\ \hline 
a nest with eggs displaying brood parasitism by a cowbird & Context & Parasitism and Symbiosis \\ \hline 
an oxpecker on a zebra & Context & Parasitism and Symbiosis \\ \hline 
bananaquit pollinating flower & Context & Parasitism and Symbiosis \\ \hline 
bird perched on a hippo & Context & Parasitism and Symbiosis \\ \hline 
giant resin bee feeding on sunflower & Context & Parasitism and Symbiosis \\ \hline 
lorikeet pollinating flower & Context & Parasitism and Symbiosis \\ \hline 
Death cap mushroom & Species & Species ID \\ \hline 
Dodder or dodder laurel & Species & Species ID \\ \hline 
Green Shore Crab & Species & Species ID \\ \hline 
Green and black poison dart frog & Species & Species ID \\ \hline 
Japanese knotweed & Species & Species ID \\ \hline 
Parasitic Honey Mushrooms & Species & Species ID \\ \hline 
Spotted Lanternfly & Species & Species ID \\ \hline 
Sunflower Sea Star & Species & Species ID \\ \hline 
Zebra Mussel & Species & Species ID \\ \hline 
a rosy wolfsnail & Species & Species ID \\ \hline 
blue dragon nudibranch & Species & Species ID \\ \hline 
bridal veil stinkhorn mushroom & Species & Species ID \\ \hline 
close-up of shagbark hickory tree bark & Species & Species ID \\ \hline 
close-up of silver maple leaf & Species & Species ID \\ \hline 
close-up of sugar maple leaf & Species & Species ID \\ \hline 
close-up of sweet cherry tree bark & Species & Species ID \\ \hline 
cross orbweaver & Species & Species ID \\ \hline 
kahili ginger plant with open fruit capsules showing seeds & Species & Species ID \\ \hline 
\end{longtable}

}

\newpage
\newcommand{\sectioncolor}{black}
\section{Datasheet}

\subsection{Motivation}

\textbf{For what purpose was the dataset created?
}
Was there a specific task in mind? Was there
a specific gap that needed to be filled? Please provide a description.
\begin{itemize}
    \item The purpose of INQUIRE is to provide a challenging benchmark for text-to-image retrieval on natural world images. Prior retrieval datasets are small and do not possess a challenge for existing models, with many being adaptations of captioning datasets. These datasets also have exactly one positive match for each query, which differs significantly from real-world retrieval scenarios where many images can be matches. The initial release of INQUIRE includes 250 queries comprehensively labeled over a pool of five million natural world images. For more information see Section~\ref{sec:benchmark_description}.
\end{itemize}

\textcolor{\sectioncolor}{\textbf{Who created this dataset (e.g., which team, research group) and on behalf
of which entity (e.g., company, institution, organization)?}
}
\begin{itemize}
    \item INQUIRE and iNat24 were created by a group of researchers from the following affiliations: iNaturalist, the Massachusetts Institute of Technology, University College London, University of Edinburgh, and University of Massachusetts Amherst. The dataset was created from data made publicly available by the citizen science platform iNaturalist~\cite{iNaturalist}. 
\end{itemize}

\textcolor{\sectioncolor}{\textbf{What support was needed to make this dataset?}
(e.g.who funded the creation of the dataset? If there is an associated
grant, provide the name of the grantor and the grant name and number, or if
it was supported by a company or government agency, give those details.)
} 
\begin{itemize}
    \item Funding for annotation was provided by the Generative AI Laboratory (GAIL) at the University of Edinburgh. In addition, team members were supported in part by the Global Center on AI and Biodiversity Change (NSF OISE-2330423 and NSERC 585136) and the Biome Health Project funded by WWF-UK. 
\end{itemize}

\textcolor{\sectioncolor}{\textbf{Any other comments?
}} 
\begin{itemize}
    \item N/A
\end{itemize}

\subsection{Composition}

\textcolor{\sectioncolor}{\textbf{What do the instances that comprise the dataset represent (e.g., documents,
photos, people, countries)?
}
Are there multiple types of instances (e.g., movies, users, and ratings;
people and interactions between them; nodes and edges)? Please provide a
description.
} 
\begin{itemize}
    \item  The dataset consists of images depicting natural world phenomena (\ie plant and animals species). In addition, it also contains natural language text queries representing scientific questions of interest. Each query is associated with a set of relevant images which came up after comprehensive labeling among the natural world image collection.
\end{itemize}

\textcolor{\sectioncolor}{\textbf{How many instances are there in total (of each type, if appropriate)?
}
}
\begin{itemize}
    \item \inquire{} contains 250 text queries and a total of 32,696 relevant image matches.
    \item iNat24 contains 4,813,543 images from 9,959 species categories.
\end{itemize}

\textcolor{\sectioncolor}{\textbf{Does the dataset contain all possible instances or is it a sample (not
necessarily random) of instances from a larger set?
}
If the dataset is a sample, then what is the larger set? Is the sample
representative of the larger set (e.g., geographic coverage)? If so, please
describe how this representativeness was validated/verified. If it is not
representative of the larger set, please describe why not (e.g., to cover a
more diverse range of instances, because instances were withheld or
unavailable).
} 
\begin{itemize}
    \item The dataset contains approximately five million images sourced from iNaturalist. This is a subset of the total number of images present on iNaturalist. The selection and filtering process used to construct the dataset is described in Section~\ref{appendix-data-annotation}.
\end{itemize}

\textcolor{\sectioncolor}{\textbf{What data does each instance consist of?
}
“Raw” data (e.g., unprocessed text or images) or features? In either case,
please provide a description.
}
\begin{itemize}
    \item Each \inquire{} instance consists of a text query and a set of images representing all relevant matches for the query within iNat24.
    \item Each iNat24 instance is an image that is associated with a set of metadata, including the species label, location (latitude and longitude), observation time, license, image dimensions, and full taxonomic classification.
\end{itemize}

\textcolor{\sectioncolor}{\textbf{Is there a label or target associated with each instance?
}
If so, please provide a description.
}
\begin{itemize}
    \item In \inquire{} each query is paired with a set of positive image matches from iNat24.
    \item iNat24 has species labels associated with each image. The species labels are obtained from `research grade' labels that have been generated from the community consensus on iNaturalist.  
\end{itemize}

\textcolor{\sectioncolor}{\textbf{Is any information missing from individual instances?
}
If so, please provide a description, explaining why this information is
missing (e.g., because it was unavailable). This does not include
intentionally removed information, but might include, e.g., redacted text.
}
\begin{itemize}
    \item There is no information relevant to the task of the dataset omitted. 
\end{itemize}

\textcolor{\sectioncolor}{\textbf{Are relationships between individual instances made explicit (e.g., users’
movie ratings, social network links)?
}
If so, please describe how these relationships are made explicit.
} 
\begin{itemize}
    \item The image id, species taxonomy, locations, and time captured are provided with each image.  
\end{itemize}

\textcolor{\sectioncolor}{\textbf{Are there recommended data splits (e.g., training, development/validation,
testing)?
}
If so, please provide a description of these splits, explaining the
rationale behind them.
} 
\begin{itemize}
    \item For \inquire{}, the queries and their relevant images are utilized solely for evaluation purposes within this paper and thus, there are no splits provided.
    \item The iNat24 dataset provides additional training data which can be used in conjunction with the validation and test splits from iNat21~\cite{van2021benchmarking}. More discussion of splits can be found in Section~\ref{appendix-data-annotation-curation}. 
\end{itemize}

\textcolor{\sectioncolor}{\textbf{Are there any errors, sources of noise, or redundancies in the dataset?
}
If so, please provide a description.
}
\begin{itemize}
    \item While the species labels for each image in iNat24 are generated via consensus from multiple citizen scientists, there may still be errors in the labels which our evaluation will inherit. However, this error rate is estimated to be low~\cite{loarie_2024}. 
    \item \inquire{} annotations may also contains noise in relevance scoring due to labeling error. However, we extensively labeled relevant queries to ensure this error rate is low.  
\end{itemize}

\textcolor{\sectioncolor}{\textbf{Is the dataset self-contained, or does it link to or otherwise rely on
external resources (e.g., websites, tweets, other datasets)?
}
If it links to or relies on external resources, a) are there guarantees
that they will exist, and remain constant, over time; b) are there official
archival versions of the complete dataset (i.e., including the external
resources as they existed at the time the dataset was created); c) are
there any restrictions (e.g., licenses, fees) associated with any of the
external resources that might apply to a future user? Please provide
descriptions of all external resources and any restrictions associated with
them, as well as links or other access points, as appropriate.
}
\begin{itemize}
    \item \inquire{} and iNat24 are self-contained datasets, as they include images and metadata that are directly available to download in their raw format without linking to any other external resources. 
\end{itemize}

\textcolor{\sectioncolor}{\textbf{Does the dataset contain data that might be considered confidential (e.g.,
data that is protected by legal privilege or by doctor-patient
confidentiality, data that includes the content of individuals’ non-public
communications)?}
If so, please provide a description.
}
\begin{itemize}
    \item No, our dataset does not contain confidential data. The images that are part of it have been made publicly available by the users of iNaturalist. 
\end{itemize}

\textcolor{\sectioncolor}{\textbf{Does the dataset contain data that, if viewed directly, might be offensive,
insulting, threatening, or might otherwise cause anxiety?}
If so, please describe why.
}
\begin{itemize}
    \item iNat24 contains pictures of the natural world (\eg plant and animal species) captured by community volunteers. Some natural world images in this dataset could be disturbing to some viewers, \eg there are a small number of images that contain dead animals. We include these images in the dataset as they are ecologically and scientifically useful, \eg for studying the impact of roadkill on animal populations.
\end{itemize}

\textcolor{\sectioncolor}{\textbf{Does the dataset relate to people?}
If not, you may skip the remaining questions in this section.
}
\begin{itemize}
    \item No, our dataset does not relate directly to people. Images of humans where their faces are visible have been filtered out using a combined manual and automated process. See Section~\ref{appendix-data-annotation-curation} for a discussion of data filtering. 
\end{itemize}

\textcolor{\sectioncolor}{\textbf{Does the dataset identify any subpopulations (e.g., by age, gender)?}
If so, please describe how these subpopulations are identified and provide a description of their respective distributions within the dataset.
} 
\begin{itemize}
    \item No, our dataset does not identify any human subpopulations.
\end{itemize}

\textcolor{\sectioncolor}{\textbf{Is it possible to identify individuals (i.e., one or more natural persons),
either directly or indirectly (i.e., in combination with other data) from the dataset?}
If so, please describe how.
} 
\begin{itemize}
    \item Some images publicly uploaded by users to the iNaturalist platform contain identifiable information, including pictures containing human faces, IDs, or license plates. To address this, we filter iNat24 to remove all such instances that we can identify, including by running detection algorithms to find all instances of human faces. More details are provided in Section~\ref{appendix-data-annotation-curation}.
    \item All photos used to construct iNat24 come from observations captured by community volunteers who have given their images a suitable license for research use. We respect these licenses by providing the license information for each image as well as the rights holder in the metadata. The user-provided rights holder name can contain the user's iNaturalist user ID. This information is already available publicly from the iNaturalist platform. 
\end{itemize}

\textcolor{\sectioncolor}{\textbf{Does the dataset contain data that might be considered sensitive in any way
(e.g., data that reveals racial or ethnic origins, sexual orientations,
religious beliefs, political opinions or union memberships, or locations;
financial or health data; biometric or genetic data; forms of government
identification, such as social security numbers; criminal history)?
}
If so, please provide a description.
} 
\begin{itemize}
\item No, our dataset does not aim to contain any data that can be considered sensitive in the ways discussed above. Details on how we filter iNat24 to remove all sensitive data are provided in Section~\ref{appendix-data-annotation-curation}.
\end{itemize}

\textcolor{\sectioncolor}{\textbf{Any other comments?
}}
\begin{itemize}
\item N/A
\end{itemize}

\subsection{Collection}

\textcolor{\sectioncolor}{\textbf{How was the data associated with each instance acquired?
}
Was the data directly observable (e.g., raw text, movie ratings),
reported by subjects (e.g., survey responses), or indirectly
inferred/derived from other data (e.g., part-of-speech tags, model-based
guesses for age or language)? If data was reported by subjects or
indirectly inferred/derived from other data, was the data
validated/verified? If so, please describe how.
} 
\begin{itemize}
    \item The queries contained within \inquire{} come from discussions and interviews with a range of experts including ecologists, biologists, ornithologists, entomologists, oceanographers, and forestry experts. This resulted in 250 text queries. 
    Annotators were instructed to label  candidate images from iNat24 as either \textit{relevant} (\ie positive match) or \textit{not relevant} (\ie negative match) to a query, and to mark an image as not relevant if there was reasonable doubt. To allow for comprehensive labeling, where applicable, iNat24 species labels were used to narrow down the search to a sufficiently small size to label all relevant images for the query of interest. The annotation process is outlined in Section~\ref{appendix-data-annotation-general}. 
    
\end{itemize}

\textcolor{\sectioncolor}{\textbf{Over what timeframe was the data collected?
}
Does this timeframe match the creation timeframe of the data associated
with the instances (e.g., recent crawl of old news articles)? If not,
please describe the timeframe in which the data associated with the
instances was created. Finally, list when the dataset was first published.
} 
\begin{itemize}
    \item The collection of iNat24 started with a iNaturalist observation database export generated on 2023-12-30. From this export, we filter observations to only include those added to iNaturalist in the years 2021, 2022, or 2023. %
    \item The collection of \inquire{} queries and comprehensive labeling of their relevant images within iNat24 took place between January 2024 (following data export from iNaturalist) and end of May 2024. 
    \item The dataset is not yet public, but will be made available prior to the NeurIPS 2024 conference conditioned on acceptance. 
\end{itemize}

\textcolor{\sectioncolor}{\textbf{What mechanisms or procedures were used to collect the data (e.g., hardware
apparatus or sensor, manual human curation, software program, software
API)?
}
How were these mechanisms or procedures validated?
} 
\begin{itemize}
\item The iNat24 dataset was sourced from a GBIF export of the iNaturalist database.
\item To comprehensively label the images that match each query in \inquire{}, we utilized a custom interface. For more information see Section~\ref{appendix-data-annotation-general}.
\end{itemize}

\textcolor{\sectioncolor}{\textbf{What was the resource cost of collecting the data?
}
} 
\begin{itemize}
\item N/A
\end{itemize}

\textcolor{\sectioncolor}{\textbf{If the dataset is a sample from a larger set, what was the sampling
strategy (e.g., deterministic, probabilistic with specific sampling
probabilities)?
}
} 
\begin{itemize}
    \item iNat24 was sampled from an export of the iNaturalist platform and consists of image observations made in the years 2021, 2022, or 2023. Details about the sampling strategy can be found in Section~\ref{appendix-data-annotation-curation}.
\end{itemize}
\textcolor{\sectioncolor}{\textbf{Who was involved in the data collection process (e.g., students,
crowdworkers, contractors) and how were they compensated (e.g., how much
were crowdworkers paid)?
}
} 
\begin{itemize}
\item 
The queries contained within \inquire{} came from discussions and interviews with a range of experts including ecologists, biologists, ornithologists, entomologists, oceanographers, and forestry experts. 
Image annotation was performed by a carefully selected team of paid MSc students or equivalent, many with expertise in ecology allowing for labeling of difficult queries. These annotators were paid at the equivalent of \$15.50 per hour.
\end{itemize}

\textcolor{\sectioncolor}{\textbf{Were any ethical review processes conducted (e.g., by an institutional
review board)?
}
If so, please provide a description of these review processes, including
the outcomes, as well as a link or other access point to any supporting
documentation.
} 
\begin{itemize}
\item 
We received internal ethical approval for our query collection and data labeling (Edinburgh Informatics Ethics Review Panel 951781 and MIT Committee on the Use of Humans as Experimental Subjects Protocol 2404001276). 
\end{itemize}

\textcolor{\sectioncolor}{\textbf{Does the dataset relate to people?
}
If not, you may skip the remainder of the questions in this section.
} 
\begin{itemize}
\item N/A. The dataset contains images of different plant and animal species that have been made publicly available by users of the citizen science platform iNaturalist under a creative commons or similar license. 
\end{itemize}

\textcolor{\sectioncolor}{\textbf{Did you collect the data from the individuals in question directly, or
obtain it via third parties or other sources (e.g., websites)?
}
} 
\begin{itemize}
\item N/A
\end{itemize}

\textcolor{\sectioncolor}{\textbf{Were the individuals in question notified about the data collection?
}
If so, please describe (or show with screenshots or other information) how
notice was provided, and provide a link or other access point to, or
otherwise reproduce, the exact language of the notification itself.
} 
\begin{itemize}
\item N/A
\end{itemize}

\textcolor{\sectioncolor}{\textbf{Did the individuals in question consent to the collection and use of their
data?
}
If so, please describe (or show with screenshots or other information) how
consent was requested and provided, and provide a link or other access
point to, or otherwise reproduce, the exact language to which the
individuals consented.
} 
\begin{itemize}
\item N/A
\end{itemize}

\textcolor{\sectioncolor}{\textbf{If consent was obtained, were the consenting individuals provided with a
mechanism to revoke their consent in the future or for certain uses?
}
 If so, please provide a description, as well as a link or other access
 point to the mechanism (if appropriate)
} 
\begin{itemize}
\item N/A
\end{itemize}

\textcolor{\sectioncolor}{\textbf{Has an analysis of the potential impact of the dataset and its use on data
subjects (e.g., a data protection impact analysis)been conducted?
}
If so, please provide a description of this analysis, including the
outcomes, as well as a link or other access point to any supporting
documentation.
} 
\begin{itemize}
\item N/A
\end{itemize}

\textcolor{\sectioncolor}{\textbf{Any other comments?
}} 
\begin{itemize}
\item N/A
\end{itemize}

\subsection{PREPROCESSING / CLEANING / LABELING}

\textcolor{\sectioncolor}{\textbf{Was any preprocessing/cleaning/labeling of the data
done(e.g.,discretization or bucketing, tokenization, part-of-speech
tagging, SIFT feature extraction, removal of instances, processing of
missing values)?
}
If so, please provide a description. If not, you may skip the remainder of
the questions in this section.
}
\begin{itemize}
    \item Besides resizing, we do not modify the images. Data cleaning is done to remove personally identifiable information or otherwise unsuitable images.
\end{itemize}

\textcolor{\sectioncolor}{\textbf{Was the “raw” data saved in addition to the preprocessed/cleaned/labeled
data (e.g., to support unanticipated future uses)?
}
If so, please provide a link or other access point to the “raw” data.
} 
\begin{itemize}
    \item N/A
\end{itemize}

\textcolor{\sectioncolor}{\textbf{Is the software used to preprocess/clean/label the instances available?
}
If so, please provide a link or other access point.
} 
\begin{itemize}
    \item We use the following to aid in preprocessing: \\
    - img2dataset: \textcolor{blue}{\url{https://github.com/rom1504/img2dataset}} \\
    - OpenCLIP: \textcolor{blue}{\url{https://github.com/mlfoundations/open_clip}} \\
    - Face detector: \textcolor{blue}{\url{https://github.com/biubug6/Pytorch_Retinaface}} 
\end{itemize}

\textcolor{\sectioncolor}{\textbf{Any other comments?
}} 
\begin{itemize}
    \item N/A
\end{itemize}

\subsection{USES}

\textcolor{\sectioncolor}{\textbf{Has the dataset been used for any tasks already?
}
If so, please provide a description.
} 
\begin{itemize}
    \item In our paper we use the \inquire{} dataset  to benchmark several multimodal models on text-to-image retrieval. It has not been used for any tasks prior to this. 
\end{itemize}

\textcolor{\sectioncolor}{\textbf{Is there a repository that links to any or all papers or systems that use the dataset?
}
If so, please provide a link or other access point.
} 
\begin{itemize}
    \item Currently there is no such repository as the dataset is not public. We will collate one after the dataset has been released. 
\end{itemize}

\textcolor{\sectioncolor}{\textbf{What (other) tasks could the dataset be used for?
}
} 
\begin{itemize}
    \item The iNat24 dataset could be used for training supervised fine-grained image classifiers. It could also be used for training self-supervised methods. The text pairs in \inquire{} could potentially be used to fine-tune fine-grained image generation models and vision language models. 
\end{itemize}

\textcolor{\sectioncolor}{\textbf{Is there anything about the composition of the dataset or the way it was
collected and preprocessed/cleaned/labeled that might impact future uses?
}
For example, is there anything that a future user might need to know to
avoid uses that could result in unfair treatment of individuals or groups
(e.g., stereotyping, quality of service issues) or other undesirable harms
(e.g., financial harms, legal risks) If so, please provide a description.
Is there anything a future user could do to mitigate these undesirable
harms?
} 
\begin{itemize}
    \item The images from the iNat24 dataset are not uniformly distributed across the globe (see Figure~\ref{fig:appendix-map}). Their spatial distribution reflects the spatial biases present in the iNaturalist platform. As a result, image classifiers trained on these models may preform worse on images from currently underrepresented regions. 
    \item To decrease this bias we sample from spatio-temporal clusters of ``observations groups''.  Observation groups are formed by grouping observations together if they are observed on the same day within 10km of each other, regardless of the observer. When sampling observations for a species, we cluster their associated observation groups using a spatio-temporal distance metric and then sample one observation per cluster in a round-robin fashion until we hit a desired sample size. When sampling within a cluster, we prioritize novel observation groups and novel users. 
    \item In future, this issue could be further mitigated as more data from currently underrepresented regions becomes available. 
\end{itemize}

\textcolor{\sectioncolor}{\textbf{Are there tasks for which the dataset should not be used?
}
If so, please provide a description.
} 
\begin{itemize}
    \item There could be unintended negative consequences if conservation assessments were made based on the predictions from biased or inaccurate models developed to perform well on \inquire{}. 
    Where relevant, we have attempted to flag these performance deficiencies in the main paper. 
    \item While we have filtered out personally identifiable information from our images, the retrieval paradigm allows for free-form text search. In real-world text-to-image retrieval applications care should be taken to ensure that appropriate text filters are in-place to prevent inaccurate or hurtful associations being made between user queries and images of wildlife. 
\end{itemize}

\textcolor{\sectioncolor}{\textbf{Any other comments?
}} 
\begin{itemize}
    \item N/A
\end{itemize}

\subsection{DISTRIBUTION}
\textcolor{\sectioncolor}{\textbf{Will the dataset be distributed to third parties outside of the entity
(e.g., company, institution, organization) on behalf of which the dataset
was created?
}
If so, please provide a description.
} 
\begin{itemize}
    \item Yes, INQUIRE and iNat24 will be publicly available for download.
\end{itemize}

\textcolor{\sectioncolor}{\textbf{How will the dataset will be distributed (e.g., tarball on website, API,
GitHub)?
}
Does the dataset have a digital object identifier (DOI)?
} 
\begin{itemize}
    \item The dataset is distributed as a tarball. Links to the dataset download are available on our GitHub repository at \textcolor{blue}{\url{https://github.com/inquire-benchmark/INQUIRE}}
\end{itemize}

\textcolor{\sectioncolor}{\textbf{When will the dataset be distributed?
}
} 
\begin{itemize}
    \item The dataset will be publicly released conditioned on acceptance.  %
\end{itemize}

\textcolor{\sectioncolor}{\textbf{Will the dataset be distributed under a copyright or other intellectual
property (IP) license, and/or under applicable terms of use (ToU)?
}
If so, please describe this license and/or ToU, and provide a link or other
access point to, or otherwise reproduce, any relevant licensing terms or
ToU, as well as any fees associated with these restrictions.
} 
\begin{itemize}
    \item The dataset will have the following ToU:  By downloading this dataset you agree to the following terms
    \begin{itemize}
        \item You will abide by the iNaturalist Terms of Service \textcolor{blue}{\url{https://www.inaturalist.org/pages/terms}}.
        \item You will use the data only for non-commercial research and educational purposes.
        \item You will NOT distribute the dataset images.
        \item The University of Massachusetts Amherst makes no representations or warranties regarding the data, including but not limited to warranties of non-infringement or fitness for a particular purpose.
        \item You accept full responsibility for your use of the data and shall defend and indemnify the University of Massachusetts Amherst, including its employees, officers and agents, against any and all claims arising from your use of the data, including but not limited to your use of any copies of copyrighted images that you may create from the data.
    \end{itemize}
\end{itemize}

\textcolor{\sectioncolor}{\textbf{Have any third parties imposed IP-based or other restrictions on the data
associated with the instances?
}
If so, please describe these restrictions, and provide a link or other
access point to, or otherwise reproduce, any relevant licensing terms, as
well as any fees associated with these restrictions.
} 
\begin{itemize}
    \item Each image is accompanied with a specific license selected by the contributor. See the dataset for details. 
\end{itemize}

\textcolor{\sectioncolor}{\textbf{Do any export controls or other regulatory restrictions apply to the dataset or to individual instances?
}
If so, please describe these restrictions, and provide a link or other
access point to, or otherwise reproduce, any supporting documentation.
} 
\begin{itemize}
    \item N/A 
\end{itemize}

\textcolor{\sectioncolor}{\textbf{Any other comments?
}} 
\begin{itemize}
    \item N/A
\end{itemize}

\subsection{MAINTENANCE}

\textcolor{\sectioncolor}{\textbf{Who is supporting/hosting/maintaining the dataset?
}
} 
\begin{itemize}
    \item The dataset is hosted on AWS supported by the AWS Open Data Program.
\end{itemize}

\textcolor{\sectioncolor}{\textbf{How can the owner/curator/manager of the dataset be contacted (e.g., email
address)?
}
}
\begin{itemize}
    \item Questions, clarifications, and issues can be raised via the GitHub page: \textcolor{blue}{\url{https://github.com/inquire-benchmark/INQUIRE}}
\end{itemize}

\textcolor{\sectioncolor}{\textbf{Is there an erratum?
}
If so, please provide a link or other access point.
} 
\begin{itemize}
    \item Issues can be raised via the GitHub page: \textcolor{blue}{\url{https://github.com/inquire-benchmark/INQUIRE}}
\end{itemize}

\textcolor{\sectioncolor}{\textbf{Will the dataset be updated (e.g., to correct labeling errors, add new
instances, delete instances)?
}
If so, please describe how often, by whom, and how updates will be
communicated to users (e.g., mailing list, GitHub)?
} 
\begin{itemize}
    \item There may be a future version of the dataset, however we do not intend for the dataset to be frequently changing. 
\end{itemize}

\textcolor{\sectioncolor}{\textbf{If the dataset relates to people, are there applicable limits on the
retention of the data associated with the instances (e.g., were individuals
in question told that their data would be retained for a fixed period of
time and then deleted)?
}
If so, please describe these limits and explain how they will be enforced.
} 
\begin{itemize}
    \item N/A 
\end{itemize}

\textcolor{\sectioncolor}{\textbf{Will older versions of the dataset continue to be
supported/hosted/maintained?
}
If so, please describe how. If not, please describe how its obsolescence
will be communicated to users.
} 
\begin{itemize}
    \item Previous versions of the iNaturalist image datasets can be found here \textcolor{blue}{\url{https://github.com/visipedia/inat_comp/tree/master}}
\end{itemize}

\textcolor{\sectioncolor}{\textbf{If others want to extend/augment/build on/contribute to the dataset, is
there a mechanism for them to do so?
}
If so, please provide a description. Will these contributions be
validated/verified? If so, please describe how. If not, why not? Is there a
process for communicating/distributing these contributions to other users?
If so, please provide a description.
} 
\begin{itemize}
    \item Contributors can join the iNaturalist platform: \textcolor{blue}{\url{https://www.inaturalist.org/}} 
\end{itemize}

\textcolor{\sectioncolor}{\textbf{Any other comments?
}} 
\begin{itemize}
    \item N/A 
\end{itemize}

\end{document}